\documentclass[examplefnt,biber]{nowfnt} % creates the journal version, needs biber version
\usepackage[utf8]{inputenc}
\usepackage{amsfonts}
\usepackage{gensymb}
\usepackage{mathtools}
\usepackage{multirow}
\usepackage{subfigure}
\usepackage{rotating}
\usepackage{longtable}
\usepackage{pdflscape}
\DeclareUnicodeCharacter{0301}{\'{e}}
\usepackage{amsmath}
\DeclareMathOperator*{\argmax}{arg\,max}
\DeclareMathOperator*{\argmin}{arg\,min}

\DeclarePairedDelimiter{\ceil}{\lceil}{\rceil}
\DeclarePairedDelimiter\floor{\lfloor}{\rfloor}

\newcommand{\ie}{\emph{i.e.}}
\newcommand{\eg}{\emph{e.g.}}

\newcommand{\NEW}[1]{{\color{black}{#1}}}

\title{A Survey of Decision-Theoretic Approaches for Robotic Environmental Monitoring}

\subtitle{Near-Earth Sensing with Robots}

\maintitleauthorlist{
Yoonchang Sung \\
\and
Zhiang Chen \\
\and
Jnaneshwar Das \\
\and
Pratap Tokekar
}

\issuesetup
{%
 copyrightowner={A.~Heezemans and M.~Casey},
 volume        = xx,
 issue         = xx,
 pubyear       = 2023,
 isbn          = xxx-x-xxxxx-xxx-x,
 eisbn         = xxx-x-xxxxx-xxx-x,
 doi           = 10.1561/XXXXXXXXX,
 firstpage     = 1, %Explain
 lastpage      = 18
 }

\usepackage{mwe}

\author[1]{Sung,Yoonchang}
\author[2]{Chen, Zhiang}
\author[3]{Das,Jnaneshwar}
\author[4]{Tokekar,Pratap}

\affil[1]{Department of Computer Science, The University of Texas at Austin, Austin, TX 78712, USA; yooncs8@cs.utexas.edu}
\affil[2]{Division of Geological and Planetary Sciences, California Institute of Technology, Pasadena, CA 91125, USA; zchen256@caltech.edu}
\affil[3]{School of Earth and Space Exploration, Arizona State University, Tempe, AZ 85287, USA; jdas5@asu.edu}
\affil[4]{Department of Computer Science, University of Maryland, College Park, MD 20742, USA; tokekar@umd.edu}

\articledatabox{\nowfntstandardcitation}

\begin{document}

\makeabstracttitle

\begin{abstract}
Robotics has dramatically increased our ability to gather data about our environments, creating an opportunity for the robotics and algorithms communities to collaborate on novel solutions to environmental monitoring problems. To understand a taxonomy of problems and methods in this realm, we present the first comprehensive survey of decision-theoretic approaches that enable efficient sampling of various environmental processes. We investigate representations for different environments, followed by a discussion of using these presentations to solve tasks of interest, such as learning, localization, and monitoring. To efficiently implement the tasks, decision-theoretic optimization algorithms consider:  (1) where to take measurements from, (2) which tasks to be assigned, (3) what samples to collect, (4) when to collect samples, (5) how to learn environment; and (6) who to communicate. Finally, we summarize our study and present the challenges and opportunities in robotic environmental monitoring. 

\end{abstract}

\chapter{Introduction}
\label{chap:intro} % a label for the chapter, to refer to it later

Environmental monitoring is a crucial field encompassing diverse applications, including marine exploration, wildlife conservation, ecosystem assessment, and air quality monitoring. Collecting accurate and timely data from inaccessible locations and challenging environments is essential for understanding and addressing environmental issues. Robots offer a promising solution by enabling data collection at unprecedented spatio-temporal scales. However, relying solely on teleoperation is impractical and limits the efficiency and effectiveness of environmental monitoring efforts. Autonomy plays a pivotal role in unlocking the full potential of robots, allowing them to operate independently and intelligently in complex environments.

This survey paper focuses on high-level decision-making problems in autonomous environmental monitoring robots. Decision-making at the high level involves strategic planning and coordination to optimize data collection. Addressing these challenges allows robots to autonomously navigate, explore, and gather scientific data in a wide range of environmental monitoring applications.

Despite the potential benefits of autonomous environmental monitoring, several research challenges must be overcome. The first challenge lies in the development of effective high-level decision-making algorithms capable of handling environmental complexities and uncertainties within resource constraints. These algorithms enable robots to make informed decisions on task prioritization, sensor selection, path planning, and collaboration with other robots or human operators. Additionally, ensuring the robustness, adaptability, and scalability of decision-making systems is critical and challenging for real-world deployments. This survey delves into the current state-of-the-art decision-making algorithms, compares their strengths and limitations, and discusses their applicability to environmental monitoring, aiming to shed light on the progress made and highlight the open research problems in this field.

By focusing on high-level decision-making, this survey aims to provide insights and understanding for researchers and practitioners in the field of autonomous environmental monitoring robotics. The knowledge gained from this survey can guide the development of advanced decision-making techniques, paving the way for more effective and efficient environmental monitoring efforts and contributing to the broader goal of sustainable resource management and conservation.

Environmental monitoring encompasses a wide range of applications. Despite the diversity of these applications, many decision-making problems share common characteristics. For example, robotic systems in environmental monitoring applications face complex decision-making challenges, requiring high-level planning to optimize resource utilization and data collection. Effective decision-making in these scenarios requires abstractions to model these environments and formulate efficient solutions. By identifying these commonalities, we can uncover general principles and techniques that can be adapted and applied across different environmental monitoring applications.

% This paper will present a taxonomy of decision-making algorithms, highlighting their strengths, limitations, and applicability to various environmental monitoring contexts. It will explore different approaches, such as optimization-based methods, reinforcement learning, graph-based algorithms, and more, in the context of decision-making at the high-level planning level. Moreover, the survey will also discuss the challenges associated with decision-making in resource-constrained environments and the open research problems that require further investigation. this survey paper aims to facilitate knowledge exchange, encourage collaboration, and foster advancements in the field. Through a better understanding of decision-making challenges and solutions, roboticists can contribute to the effective monitoring and conservation of our environment, ultimately serving the broader environmental monitoring community.

Particularly, we focus on three typical decision-making tasks in the scientific studies of environmental monitoring. First, environmental scientists wish to efficiently \emph{learn} representations for environmental processes. Second, they want to \emph{localize} phenomena such as hotspots using these representations. Third, they want to \emph{monitor} change in the phenomena, \eg, movement in the boundary of an oil spill and the change in tree canopy size through a season. This article presents a comprehensive survey of environmental monitoring, covering data-driven algorithms geared towards deployment on cyber-physical systems, such as wireless sensor networks and networked robotic vehicles. The paper serves as a tutorial for engineers and scientists who are interested in applying information theoretic algorithms to maximize the yield of scientific studies. This objective is particularly necessary in the context of field data collection that can be dramatically improved by the choice of efficient robotic sampling strategies.

%we need a transition paragraph that will tell the reader what this paper will be about
%We present a survey of current environmental robotics applications and decision theoretical approaches for addressing key challenges 
% {\small What are the science needs? What are the available technologies to acquire and analyze data to meet science needs? Can better algorithms be developed to improve data collection and/or analysis?} 
% \section{Applications}\label{sec:app}

% The application of robotic sampling is still in its early stages, and there is potential for wide application across various domains of earth as well as space sciences. We will use three motivating examples from our own engagements with domain scientists in the arena of aquatic and terrestrial monitoring to lay out the algorithmic landscape that can improve science yield. In the terrestrial domain, we have collaborated with earthquake scientists, drylands ecologists, and agronomists, to address questions around fracturing of rocks, microbes in extreme conditions, and optimization of crop health and yield. In the aquatic domain, we have carried out both large scale and small scale marine and littoral robotic sampling campaigns. 

\section{Related Surveys}
\label{sec:survey}
We compile related survey papers that discuss autonomous monitoring applications. \citet{corke2010environmental} review environmental and agricultural applications utilizing wireless sensor networks. \citet{camilli2010bright} conduct an overview of advanced platforms, power generation, communications, and sensing technologies for marine environmental monitoring. An environmental monitoring survey paper (\cite{dunbabin2012robots}) introduces various sensor types, sensor network platforms, and commonly used technologies for measuring environmental variables. \citet{lattanzi2017review} present a recent survey on robotic infrastructure inspection. Challenges in the infrastructure inspection include the design of inspection robots, trajectory planning, and handling GPS-denied environments. Precision agriculture (\cite{r2018research,vougioukas2019agricultural,sparrow2021robots,oliveira2021advances,basiri2022survey}) has recently gained significant attention in robotics, with an objective to monitor and enhance crop health by utilizing multiple sources in a more efficient manner. \citet{murphy2016disaster} summarize robot designs, concepts, and open issues in search and rescue for disaster scenarios. \citet{liu2013robotic} introduce control methods in robotic urban search and rescue, while a survey paper by \citet{queralta2020collaborative} focuses on multi-robot systems for search and rescue. 

In contrast to the aforementioned survey papers that focus on specific applications, our goal is to offer a comprehensive algorithmic framework for a broad spectrum of environmental monitoring tasks. Our paper will serve as a guideline for identifying appropriate decision-making challenges, assumptions, and algorithmic considerations for various applications.

\begin{figure*}
\centering
\includegraphics[width=\textwidth]{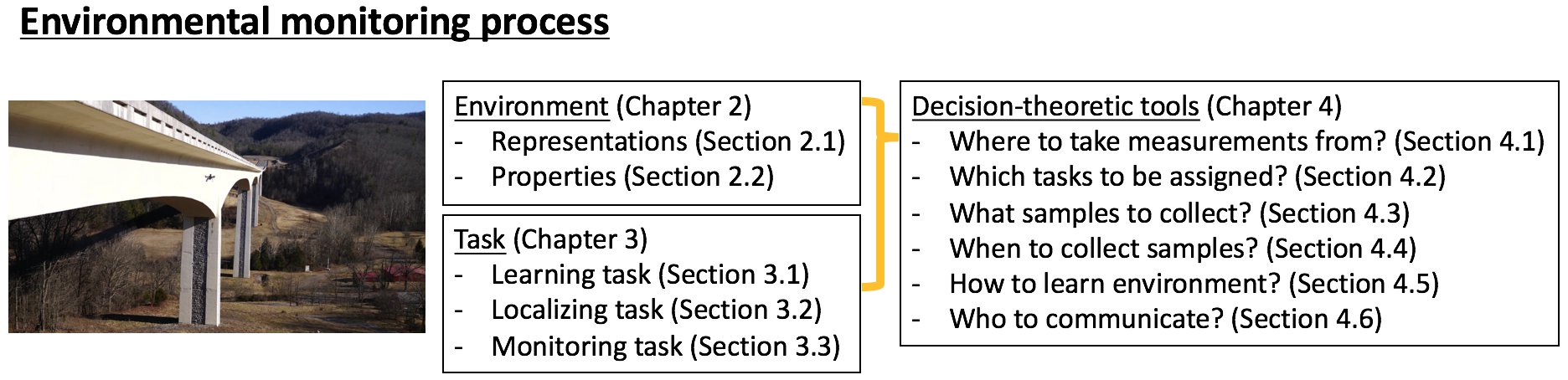}
\caption{\NEW{Organization of this survey. Environmental monitoring processes involve three key aspects that must be clearly defined. First, it is essential to determine the ideal environmental representations and their properties. Second, understanding the type of task the monitoring process is involved in is crucial to define task objectives effectively. Third, leveraging decision-theoretic tools from the literature is critical for making informed decisions during environmental monitoring. This survey provides detailed guidelines for each of these aspects to assist practical users in their monitoring endeavors. 
}}
\label{fig:organization}
\end{figure*}

The remaining of this survey is organized as follows (Figure~\ref{fig:organization}). In Chapter~\ref{chap:env}, we introduce various methods of representing the environment, which are broadly grouped into discrete and continuous approaches. In Chapter~\ref{chap:task}, we present three primary tasks in environmental monitoring: the learning, localizing, and monitoring tasks. Chapter~\ref{chap:decision} is the core of this survey, where we delve into the details of various decision-theoretic approaches based on 5W1H-driven categories. We conclude the survey in Chapter~\ref{chap:conc} with a discussion on promising future directions and final remarks.

\chapter{Robust Representations of the Environment}\label{chap:env}

The decision-making algorithms for robotic environmental monitoring depend on environmental representations. Choosing the appropriate representation requires a careful balance between the fidelity of the representation and its suitability for planning and computation. In this chapter, we survey the commonly used environmental representations and discuss their properties. 

\section{Representations}~\label{sec:rep}
Environmental representations can be broadly categorized into discrete and continuous representations. Hybrid representations, such as variables on continuous space and discrete time, are also common. Planning algorithm design and analysis depend closely on the choice of representation. 

\subsection{Discrete representations}\label{subsec:discrete}

Discrete representations are common due to their simplicity. 
Using a denser resolution for discretization (higher sampling frequency) yields results closer to the ground truth but demands greater computational resources. 

Common representations of discrete environment are \emph{grid cells} and a \emph{graph}. Grid cells partition an environment space into equal-sized cells. The shape of a grid cell is usually a square~\citep{sung2019competitive}. Other shapes, \eg, a hexagonal cell~\citep{sujit2004search}, are employed to enable more flexible robot actions. When considering a 3D environment explored by aerial robots, a quadtree structure may be suitable to accommodate the varying resolutions of sensor footprints~\citep{carpin2013variable,vidal2018optimized}. \NEW{The time complexity for accessing grid cells is constant, denoted as $\mathcal{O}(1)$. The complexity for searching is linear, $\mathcal{O}(n)$, where $n$ is the total number of grid cells. The space complexity of storing data in grid cells is $\mathcal{O}(n)$.}

\emph{Occupancy grid map} representation~\citep{thrun2003learning} is usually used to delineate free cells that robots can navigate and to identify occupied cells as obstacles. Octomap~\citep{hornung2013octomap} is a 3D probabilistic occupancy map that incorporates uncertainty induced by imperfect sensing based on an octree structure~\citep{palomeras2019autonomous}. \NEW{The time complexity for accessing, inserting, or updating a node in the octomap is $ \mathcal{O}(\log n)$, where $n$ is the number of nodes. Memory usage is adaptive, often more efficient than grid maps, especially in vast environments with large homogenous regions.} Temporal aspects have been incorporated in occupancy maps~\citep{arbuckle2002temporal} to model a spatiotemporal environment. %Recent works on continuous occupancy mapping do not require a predefined resolution of a map; we introduce these works in Section~\ref{subsec:continuous}.

\emph{Signed distance function} (SDF) computes the distance of a given point from the boundary of the nearest object where a sign determines whether the point is inside of the object or not. \citet{curless1996volumetric} propose a discretized signed distance function called signed distance fields. In comparison with occupancy grid maps, the signed distance field representation facilitates collision checking and accurately represents the contour of boundary contours ~\citep{saulnierinformation}. \NEW{The computational complexity of evaluating or computing the SDF often depends on the representation and method used. For example, in grid cells (or voxels), the SDF values are usually precomputed for every voxel in a grid. The time complexity for querying the SDF value of a point is  $\mathcal{O}(1)$. Computing or updating the SDF values for all voxels in the grid can be  $\mathcal{O}(n)$, where $n$ is the number of voxels.  }

\emph{Graph} representation is used to symbolically decompose an environment into a set of regions that robots can visit and a set of paths that connect the regions. In terms of graph theory, regions and paths correspond to nodes and edges, respectively. Edges can be either directed or undirected depending on traversability between adjacent nodes in the problem. The cost of moving along an edge can be represented by a weighted edge. The graph map representation is also called \emph{topological map}, which is reduced from a metric map to represent the relations among entities. \emph{Topometric map} is additionally involved with local metric maps connected by edges, \eg, \emph{Voronoi graph}~\citep{osswald2016speeding}. Grid cells are a special graph named a grid graph.

\NEW{
\emph{Gaussian Markov random fields} (GMRF; \cite{lindgren2011explicit}) is a type of probabilistic graph that utilizes undirected graphs to represent conditional dependencies between variables and leverage Gaussian distributions to describe the probabilistic relationships. This modeling approach harnesses the inherent spatial Markov property and has showcased significant promise in the field of robotics. Its appeal lies in its remarkable computational efficiency and scalability, making it a valuable tool in the realm of robotic systems~\citep{xu2016bayesian,kreuzer2018learning,duecker2021embedded}.
}

% (PO)MDP.
\subsection{Continuous representations}\label{subsec:continuous}
\subsubsection{Scalar environment field}
The \emph{scalar environment field} is used to represent a quantity of interest, such as temperature and salinity. Robots can measure those quantities within an $n$-dimensional environment through a continuous environmental function that maps from $\mathbb{R}^n$ to $\mathbb{R}$.

\subsubsection{Gaussian processes}
\emph{Gaussian processes} (GPs) are nonparametric Bayesian methods designed to model spatio-temporal phenomena. GPs rely on a finite collection of variables, which are Gaussian-distributed, to represent these features of interest. They generate features of interest in any continuous space and time and provide an associated prediction uncertainty. Due to its simple implementation, practical usefulness, and theoretical results (\eg, near-optimality derived using submodularity;~\cite{krause2008near,Srinivas:2010:GPO:3104322.3104451}), GP has been employed with great popularity in literature~\citep{jadaliha2013gaussian,nguyen2016information,ma2018data}. GPs have also been widely applied in multi-robot systems~\citep{chen2015gaussian,luo2019distributed,ouyang2020gaussian}. To address scalability concerns, each robot is designated to manage local information.

One challenge of employing the GP is its cubic computational complexity with respect to the number of training samples, denoted by $\mathcal{O}(n^3)$. Because $n$ is usually large in environmental monitoring applications, many approximation methods have been proposed to alleviate the complexity.\footnote{More detailed summary on GP approximations can be found by~\citet{park2020gaussian}.} The most well-known approach proposed by~\citet{snelson2006sparse} involves introducing sparse pseudo inputs, referred to as inducing points, to model smoothly-varying functions with high correlations. Their model reduces complexity to $\mathcal{O}(n_i^2n)$, where $n_i$ denotes the number of inducing points.
To model highly-varying functions with low correlations, local approximation methods by~\citet{deisenroth2015distributed} divide an environment into $\frac{n}{n_i}$ regions and apply a local GP in each region, reducing the complexity to $\mathcal{O}(n_i^2n)$.
Infinite-horizon Gaussian process (IHGP) approximation by~\citet{solin2018infinite} is another local method that builds a linear Gaussian state space model (instead of a kernel matrix) and applies a Kalman filter. The complexity of IHGP is $\mathcal{O}(d^2n)$, where $d$ is the dimension of the state space. 
\citet{park2020gaussian} improve IHGP by learning time-varying hyperparameters using a sparse and nonstationary data stream. \NEW{\citet{pillonetto2018distributed} use a subset of orthonormal eigenfunctions from the Karhunen-Loève expansion to collaboratively estimate the GP, thereby reducing computational and communication burdens while maintaining estimation accuracy.}

In contrast to discrete occupancy maps, continuous occupancy mapping generates a map at any resolution, free from the constraints of single-scale mapping, thereby eliminating pre-discretization inaccuracies. 
Discrete occupancy maps usually consider the information on each grid as independent from others. Several continuous counterparts have been proposed to alleviate those shortcomings.
\emph{Gaussian process occupancy maps}~\citep{o2012gaussian,wang2016fast} handle the situation when an underlying environment exposes a structure, by encoding dependencies between grids. 
\emph{Hilbert map}~\citep{ramos2016hilbert} is another continuous occupancy map that projects data into the Hilbert space to preserve local information, allowing for fast kernel approximation.
Dynamic environments have also been studied in Gaussian process occupancy maps~\citep{senanayake2017learning} and Hilbert maps~\citep{guizilini2019dynamic}.
\emph{Bayesian Hilbert maps}~\citep{senanayake2017bayesian} use kernels to perform Bayesian logistic
regression in a high-dimensional feature space, resulting in the cubic computational complexity with respect to the number of features, which are usually much smaller than training samples.
\emph{Automorphing Bayesian Hilbert maps}~\citep{senanayake2018automorphing} improve the accuracy over Bayesian Hilbert maps by learning all location-dependent nonstationary kernel parameters with variational inference.
As automorphing Bayesian Hilbert maps require to collect the entire dataset, \citet{tompkins2020online} propose online domain adaptation to support sequential training. 
Bayesian kernel inference-based mapping by \citet{doherty2019learning} deals with sparse and noisy sensor data issues by inferring the state of unknown regions from neighboring regions. The computational efficiency of this approach, which scales logarithmically with the size of the training data, notably surpasses that of GP-based methods.

\emph{Gaussian process implicit surfaces}~\citep{williams2006gaussian,hollinger2013active}, a nonparametric regression model for reconstructing surfaces from 3D data, are adopted by~\citet{leeb2019online} to construct continuous mapping online. The proposed map computes a probabilistic estimate of the signed distance field while preserving distance gradients, which is useful for obstacle avoidance.

\subsubsection{Modeling with basis functions}
\citet{lynch2008decentralized} propose a parameterized approximation to represent the environmental function as a weighted sum of a finite set of basis functions. Examples of basis functions include sinusoids in Fourier series, wavelets, and polynomials. 
\citet{jadaliha2013environmental} consider weights as time-varying coefficient vectors to model a time-varying scalar environment field.

\subsubsection{Partial differential equations}
A more realistic environmental model is needed to track dynamic plumes that have complex behaviors. 
\citet{fahad2015robotic} use a Lagrangian model to capture both time-averaged and instantaneous structures by assuming the plume to be composed of particles.
\citet{wang2019dynamic} present an advection-diffusion partial differential equation model to model dynamic dispersion. 

% \PRT{maybe this can be moved up to the basis functions subsection? or that subsection can be moved down here} \yoon{Moved the basis function subsection here. Please feel free to remove or comment out the comments.}

Similar to the complexities in modeling dynamic plumes, achieving precise flow representation presents significant challenges, thereby compromising its reliability in robotic control applications. Several control schemes have been proposed to correct the modeling error, especially in underwater applications, such as data-driven flow models using basis functions~\citep{chang2017motion}, incompressible flow fields~\citep{lee2019online}, and gyro-like flow fields~\citep{knizhnik2022flow}.
\citet{salam2022learning} propose a method to capture high-level features in flow-like environments where obtaining an explicit flow representation is intricate.

\emph{Sparsity} in data across various domains is leveraged by choosing a suitable feature space, such as wavelets, to create parsimonious representations. If an alternative feature space can be identified, and the signal is incoherent, a sampling scheme can be developed to utilize parsimonious representations for efficient environmental observation. This approach, called compressive sensing, has demonstrated significant promise in magnetic resonance imaging and holds potential for environmental monitoring.

\subsection{Hybrid representations}\label{subsec:hybrid}

% \PRT{I feel like DP and Topic modeling should really be in a separate section called Hybrid Representation instaed of under continuous.. since its a mix of discrete (topics) and continuous (spatial grounding). What do you think?} \yoon{I agree. I created a new subsection.}

Hybrid representations are often employed to model the combination of discrete variables (\eg, environment partitions) and continuous variables (\eg, spatial groundings) for environmental phenomena.

\emph{Dirichlet processes}~\citep{teh2010dirichlet} are another nonparametric Bayesian methods, often used for clustering problems where a cluster size is unknown. In Dirichlet processes, the number of clusters is automatically learned in a data-driven manner (\ie, unsupervised learning) as the task progresses, which is an attractive feature for environmental monitoring applications. 
Several existing methods adopt Dirichlet processes to leverage such an advance.
\citet{ouyang2014multi}   propose a method that learns to partition the environmental phenomenon into local regions based on a stationary spatial correlation structure. This method utilizes a Dirichlet process mixture of locally stationary GPs. 
% \PRT{can you mention how DP is used here?} \yoon{modified}
\citet{girdhar2019streaming} employ Chinese restaurant processes, a discrete-time stochastic process closely related to Dirichlet processes, to handle a priori unknown numbers of scene labels from a discretized environment. Chinese restaurant processes model the partition of customers into tables in a restaurant context.
% \PRT{do you mean indices} \yoon{modified} 
\citet{steinberg2011bayesian} use a variational Dirichlet process model~\citep{kurihara2006accelerated} to cluster large quantities of seafloor imagery. 

\emph{Topic modeling} developed for text analysis has been adopted to produce low-dimensional descriptors that identify semantically similar or distinct objects in the environment~\citep{girdhar2014autonomous,girdhar2016modeling}. These image descriptors, formulated through topic modeling, adeptly discern thematic shifts in scenes and are robust to low-level image changes. Observations are characterized by a distribution of spatiotemporal topics. The objective of topic modeling is to compute a surprise score for a newly observed image using the distance to the closest sample in the topic space.
Several extensions have been made to topic modeling, such as multi-robot system applications~\citep{doherty2018approximate} and a hierarchical topic model for the spatial distribution of categorical observations (also known as the Gaussian-Dirichlet random field;~\cite{soucie2020gaussian}).

\section{Properties}~\label{sec:prop}
Understanding and identifying properties of environments are important due to their large impact on algorithm design. By doing so, we can narrow down to particular algorithms from a large class of algorithm options and find useful property-based guarantees that solve a given environmental monitoring problem. We show properties of environments that might vary greatly depending on a problem instance.

\subsection{Phenomenon countability}
The target monitoring phenomena can be categorized into \emph{discrete objects} and \emph{continuous fields}. Discrete objects are countable and can be represented by points, \eg, humans, animals, rocks, and fruits. Continuous fields are uncountable and can be represented by the density of environmental attributes, such as gas, water temperature, ocean salinity, and soil nitrogen levels. A special type of continuous field forms a specific region of interest with an arbitrary shape and is often not entirely detectable by a single footprint of a robot sensor due to its large size. Examples of this variant include plumes of pollutants, radioactive emissions, and fires.

\subsection{Field (non)-stationarity}
Most environments for continuous fields vary spatially in practice. A nonstationary property usually exhibits different degrees of smoothness in the local variation in different regions of an environment~\citep{ouyang2014multi}. Nonstationary versions of the covariance functions have been proposed for GP~\citep{paciorek2004nonstationary}. 
\citet{chen2022ak} design nonstationary kernels, called the attentive kernel, for better accuracy and uncertainty quantification.
The work by \citet{cao2013multi} exploits the spatially correlated structure of GP to enhance the performance of their algorithm. They investigate an environment with higher spatial correlation along one direction than in the perpendicular direction, referred to as an \emph{anisotropic} field. 

\subsection{Temporal characteristics}
A binary environment consists of regions of interest and regions of non-interest~\citep{sung2019competitive}. This method is particularly useful for coverage and exploration problems where an objective is not to estimate the density of the environment but to completely cover every single point in regions of interest. 

Generally, the states of an environment may vary over time due to external factors, for example, natural phenomena like wind and waves. The location of discrete objects and the density of continuous fields may change temporally. To estimate time-varying states of discrete objects, conventional state estimators, such as Kalman filters~\citep{lynch2008decentralized}, extended Kalman filters~\citep{hitz2015relaxing}, and particle filters~\citep{korner2010autonomous}, are useful to incorporate noisy measurements. Searching for and tracking a single mobile target is still actively studied in challenging scenarios, such as a pursuit evasion problem~\citep{kalyanam2020graph}. As conventional state estimators are not suitable for handling a varying number of objects, \citet{mahler2003multitarget} proposes the probability hypothesis density filter that simultaneously estimates both the state and the number of objects. To avoid the expensive computation of the probability hypothesis density filter, sampling-based~\citep{dames2017detecting} and Gaussian-based~\citep{sung2021gm} approximations have been adopted. \citet{krajnik2015s} propose a frequency-based map to represent the periodicity of environment states, \eg, human daily regular activities.

Previous work has considered both discrete and continuous time aspects to handle temporal dynamics in continuous fields. In the case of discrete time, \citet{ma2018data} assume piecewise static intervals to approximate temporal dynamics. GP is often used to model spatiotemporal environments in the continuous-time domain. Most approaches that utilize GPs introduce an additional dimension to represent time, allowing the time domain to be treated independently of the state-space domain~\citep{binney2013optimizing,marchant2012bayesian}. \citet{singh2010modeling} analyze the effect of correlating time with state space in terms of a covariance function in GP (\ie, non-separability). They design various covariance functions that exhibit different combinations of stationarity and separability. Their empirical findings indicate that nonstationary separable covariance functions are the most suitable for spatiotemporal GP models.

\subsection{Measurement models}
When the environmental monitoring task is data collection---to obtain sensor measurements or collect samples, the characteristics of measurements or samples represent the environment. While previous work mostly assume a measurement noise to be Gaussian, \citet{wu2012robust} consider an unknown non-Gaussian noise and propose a control method that only requires the bounded power of noises. Sampling methods can be either \emph{in situ} or \emph{ex situ}. In situ samples are analyzed online while a robot explores an environment. The robot can adapt its precomputed plan using online streaming information extracted from in situ samples. Ex situ samples, however, are studied offline in the lab. Ex situ sampling requires a robot to plan its trajectory to collect the most valuable samples while considering its limited load capacity~\citep{das2015data,flaspohler2018near}. Lastly, heteroscedastic environments where noise properties vary in state and action spaces have been investigated by~\citet{martin2017extending}.

%\subsection{Environmental Monitoring Problem}~\label{subsec:problem}

% \begin{figure}
%     \centering
%     \includegraphics[width=3in]{fig/Environmental_Monitoring_Big_Picture.png}
%     \caption{Science questions provide the objectives and constraints that determine the experimental design and algorithmic choices, leading to low noise scientific insights.}
%     \label{fig:my_label}
% \end{figure}
%%%%%%%%%%
\chapter{Using Representations to Solve Tasks}\label{chap:task}
Having discussed the representation of the environments, in this section, we identify and describe three environmental monitoring tasks: \emph{learning}, \emph{localizing}, and \emph{monitoring}. We then present what specific problems have been addressed for each task in the literature.

\section{Learning}~\label{sec:learning}
\begin{figure}[htb]
\centering
\subfigure[A quadcopter learning the crop height field over a farm using a LIDAR sensor (reprinted from~\cite{suryan2020learning} with permission).]{\includegraphics[width=0.49\columnwidth]{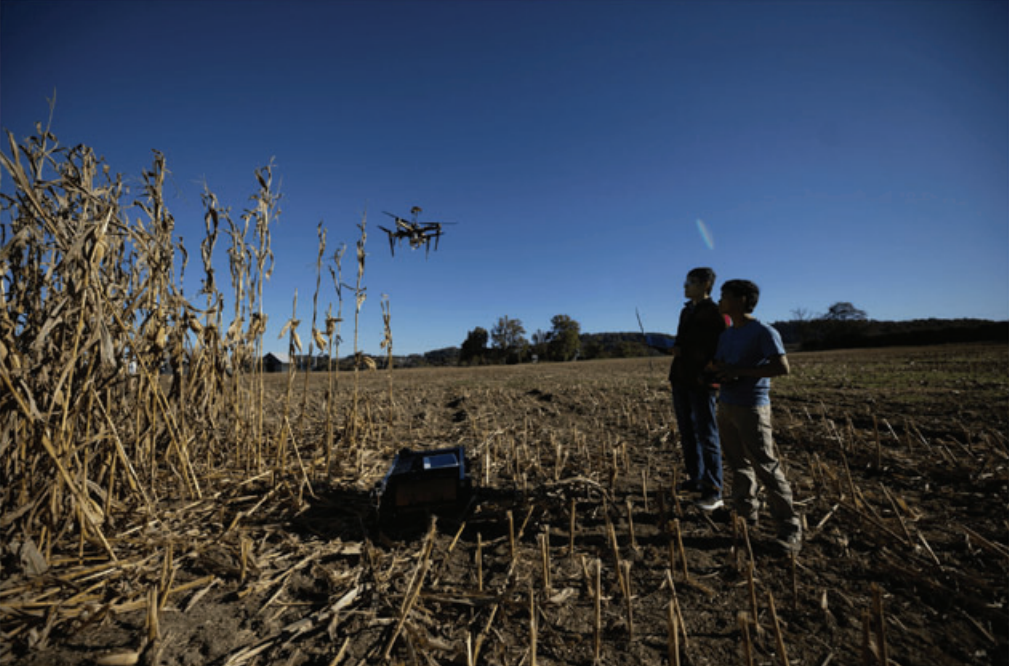}}
\subfigure[Learning spatial distributions of rock traits from aerial vehicles using a camera sensor (reprinted from~\cite{chen2020geomorphological} with permission).]{\includegraphics[width=0.50\columnwidth]{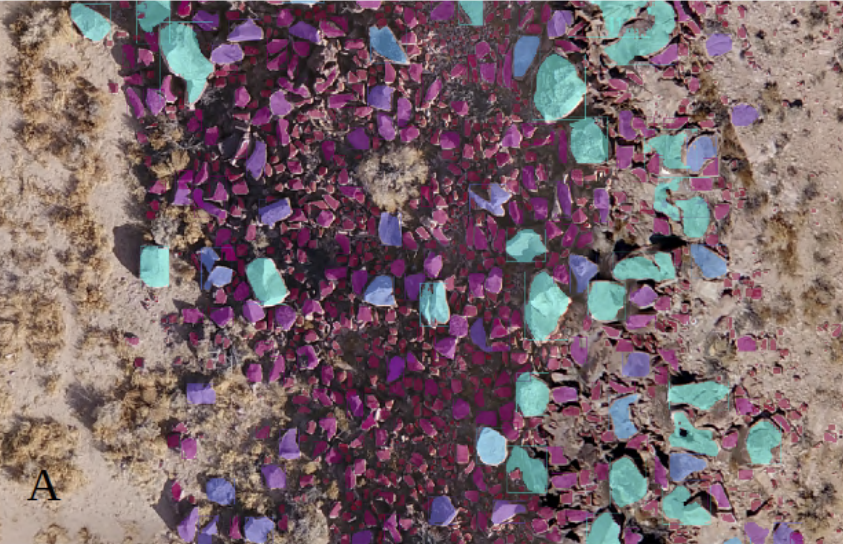}}
\caption{
Example learning tasks.
}
\label{fig:learning}
\end{figure}

Environmental monitoring involves robots exploring the environment to gather information, either measured by an onboard sensor or physically collected, to understand a domain of interest. From the local information collected, global characteristics of the environment must be inferred through certain model learned from data. Therefore, deciding where to collect sensor information or samples next is crucial. As such, learning is an important component for successful environmental monitoring (refer to the example tasks in Figure~\ref{fig:learning}).

\subsection{Accurate modeling}\label{subsec:learning-modeling}
Learning an environment model is a core challenge to carrying out downstream tasks, especially when exploring an unknown environment.
Various statistical models can be adopted for environment modeling from noisy observations.

Among statistical models, \emph{expectation maximization} (EM) is an iterative method to estimate the parameters of an environment model that depends on latent variables. EM is particularly suitable for underwater applications where the ocean current cannot be directly measured but estimated indirectly via other properties, such as position drift. 
\citet{lee2019online} investigate this problem and introduce a Gaussian process-based EM algorithm for estimating ocean currents for navigation and planning.

Accurate modeling is often intractable in practice; thus, 
approximate state estimation methods have been proposed to simplify intractable statistics. 
For example, \citet{jadaliha2013gaussian} address the challenge of resource-constrained sensors, outputting observations under localization uncertainty. The analytically intractable posterior predictive statistics are approximated using Monte Carlo sampling and Laplace's method. They analyze the approximation error and complexity (\ie, trade-offs between the error and complexity of Laplace approximations).

Environment modeling by multi-robot systems has been researched with tools such as a distributed consensus estimator.  
\citet{lynch2008decentralized} propose a decentralized, scalable approach to model the environment based on average consensus estimators.

\subsection{Mapping}\label{subsec:learning-mapping}
An alternative approach to learning about the environment is to explore and cover an environment (or a partial environment depending on given constraints). Robots with limited sensing capability gradually estimate the location, shape, and size of the environment. This process of incrementally building environment information is known as \emph{mapping}. Offline mapping techniques construct a map by utilizing all collected observations and the entire history of robot states. In contrast, online mapping techniques continuously update the map based on the latest observations and the most recent robot states.

% For instance, scene mapping by~\cite{girdhar2019streaming} focuses on developing a scene understanding model that characterizes the environment properties to aid human operators. 

\subsection{Lifelong learning}\label{subsec:learning-lifelong}
One of the desired goals for any learning method is to continually learn new behaviors and refine a model over a long period. \emph{Lifelong learning} refers to such capability. 

\citet{girdhar2016modeling} propose curiosity modeling for long-term exploration in aquatic environments, recognizing that much of the collected data contains uninteresting observations. Their objective is to achieve lifelong learning behavior in the perception model through topic modeling, guiding the robot to locations with high topic perplexity to enhance learning.

\section{Localizing}~\label{sec:localizing}
\begin{figure}[htb]
\centering
\subfigure[An aerial vehicle exploring the environment to search for the hotspot of a hazardous plume in a lake (reprinted from~\cite{sung2019multi} with permission).]{\includegraphics[width=0.59\columnwidth]{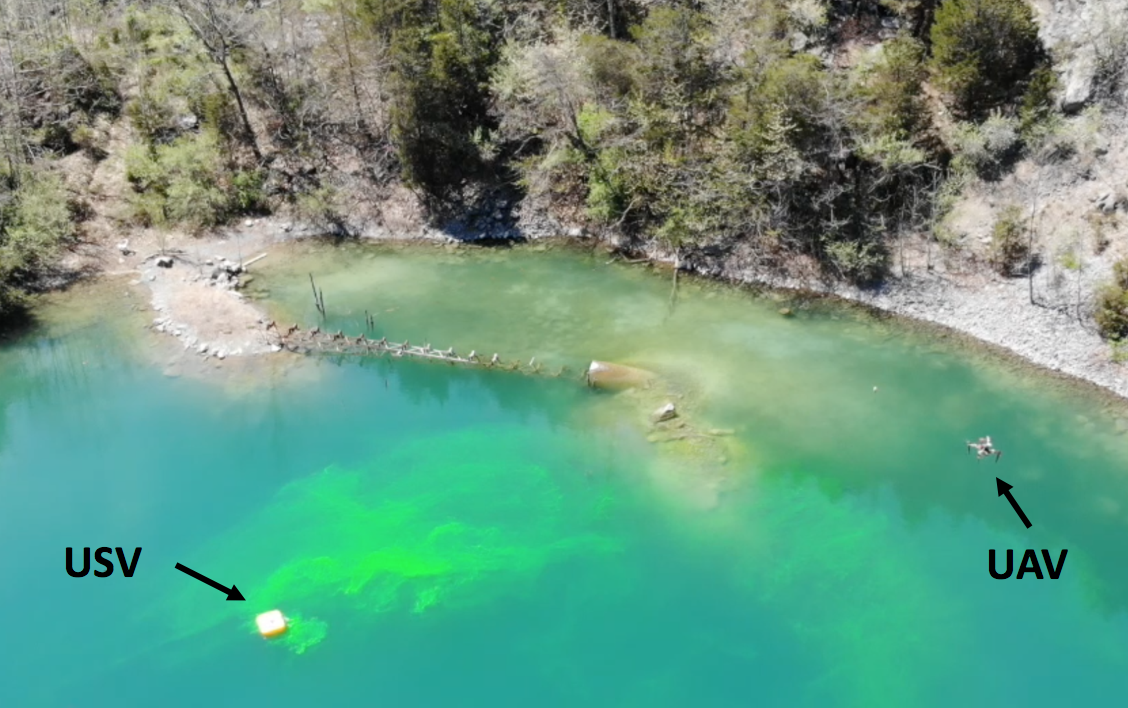}}
\subfigure[Hotspot search task for harmful algal blooms applied to underwater vehicles (reprinted from~\cite{das2010towards} with permission).]{\includegraphics[width=0.40\columnwidth]{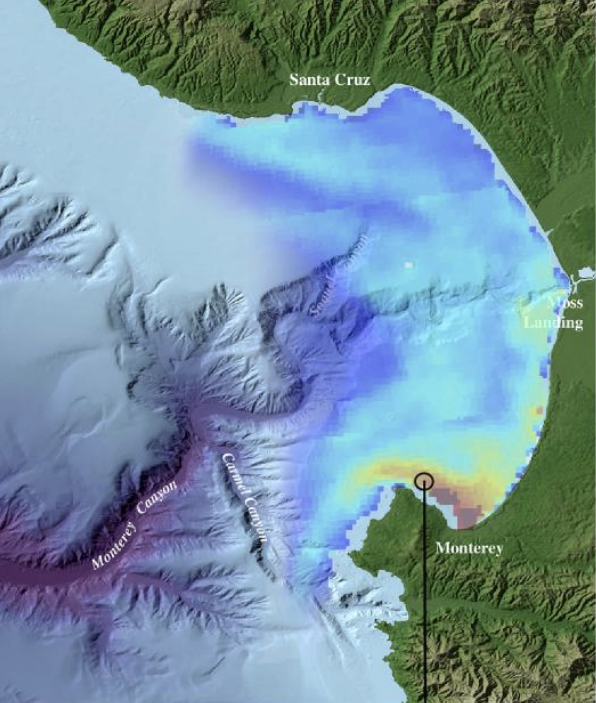}}
\caption{
Example localizing tasks.
}
\label{fig:localizing}
\end{figure}

Localizing an object of interest from a large-scale environment is undoubtedly critical in environmental monitoring (refer to the example tasks in Figure~\ref{fig:localizing}). Depending on the characteristics of an object, we categorize localization into \emph{search and tracking} for dynamic objects and \emph{hotspot identification} for stationary objects.

\subsection{Search and tracking}\label{subsec:localizing-tracking}

Common \emph{search and tracking} methods include Kalman filters, particle filters, and probability hypothesis density filters (see Section~\ref{sec:prop}). 
For example, \citet{van2019online} introduce an algorithm based on the probability hypothesis density filter to simultaneously detect and track unknown radio-tagged objects. 

Plume tracking searches for and tracks the source of a plume (\eg, radioactive dispersal and oil spill). The location, shape, and size of the plume fluctuate as it is carried along by water currents or air flow. 
\citet{neumann2013gas} tackle gas source localization using a micro-UAV and present bio-inspired plume tracking algorithms.
\citet{hajieghrary2017information} frame source seeking in plume tracking as an information-theoretic search problem, with the goal of determining actions for multi-robot systems that maximize the change in entropy.
\citet{sung2019competitive} consider an oil-spill scenario where the shape and size of the plume are unknown.

\emph{Search and rescue} is an active research field that may not be directly related to environmental monitoring. However, algorithms developed in search and rescue embrace objectives and contributions desirable for environmental monitoring. For example, localization is also a key challenge in search and rescue.

An extensive survey on search and rescue can be found in~\citet{queralta2020collaborative}.
\citet{liu2013robotic} survey search and rescue in urban applications.
Some specific objectives include maximizing search efficiency under time constraints~\citep{meera2019obstacle}, optimal deployment~\citep{macwan2011optimal}, and minimizing the mean-time-to-find and maximizing the target finding probability~\citep{meghjani2016multi}.

Harsh conditions on observing measurements have been researched, such as \citet{song2011time,song2012simultaneous,kim2014cooperative}.
\citet{song2011time} study the problem of searching for a stationary target emitting short-duration signals intermittently under limited sensing conditions. This scenario can include situations such as locating an airplane black box or finding an earthquake victim emitting signals periodically.
\citet{song2012simultaneous} introduce additional challenges compared to their previous work, including source anonymity and an unknown number of sources. To address these challenges, they employ a spatiotemporal probability occupancy grid to model the radio source and develop a motion planning algorithm to ensure that the robot traverses high-probability regions. In their subsequent work~\citep{kim2014cooperative}, they extend this approach to multiple robots, with a focus on pairs of robots.

\subsection{Hotspot identification}\label{subsec:localizing-hotspot}
Finding hotspots from the field requires efficient exploration and sampling strategies. 
\citet{low2008adaptive} propose an adaptive exploration strategy for a team of robots to simultaneously localize hotspots and learn map phenomena. 
\citet{garcia2012online} focus on spatially distributed sample collection from hotspots, involving the online calculation of sample utility. This approach can incorporate science preferences and evolving knowledge about the feature.

Radio signal mapping is one of the prime applications in environmental monitoring.
Online radio signal mapping is crucial for wireless communication to model radio signal propagation. Its primary objective is to localize the signal source in an unknown environment.
\citet{fink2010online} suggest mapping received radio signal strength in a Gaussian process using a team of mobile robots. They propose an exploration strategy based on the gradient of the predictive variance and an exploitation strategy based on maximum likelihood estimation.
\citet{atanasov2015distributed} design distributed algorithms for a team of robots to localize the source of a noisy signal. They particularly consider two scenarios: one with an available signal model and another one that is model-free.
\citet{al2018gradient} develop a distributed source-seeking strategy for robotic swarms without requiring gradient estimation or sharing measurement among robots.

Previous work has researched detecting and localizing a leak, similar to hotspot identification.
\citet{bennetts2013towards} focus on methane leak detection. Their algorithm generates 3D concentration maps using depth measurements and integral concentration to pinpoint the location of a leak.

\section{Monitoring}~\label{sec:monitoring}
\begin{figure}[htb]
\centering
\subfigure[A robotic raft monitoring common carp tagged with radio transmitters in a lake (reprinted from~\cite{tokekar2010robotic} with permission).]{\includegraphics[width=0.59\columnwidth]{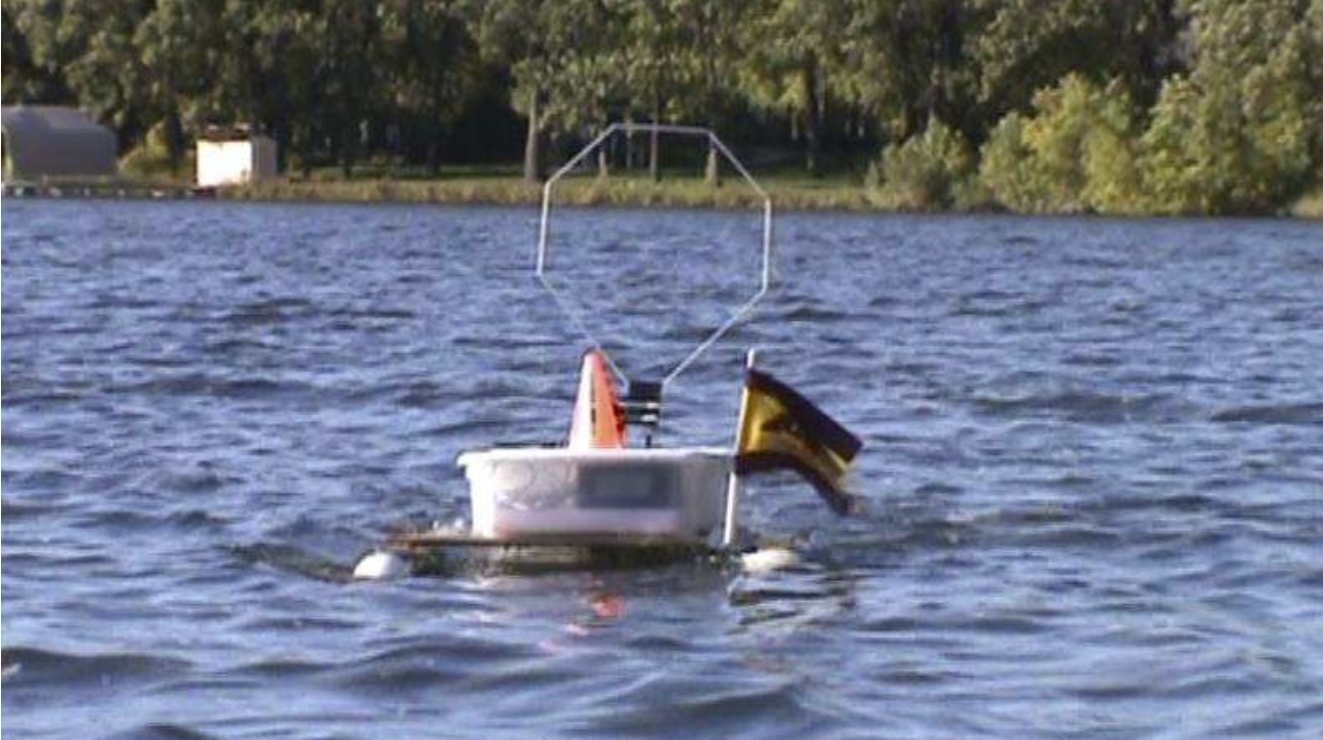}}
\subfigure[An aerial vehicle measuring soil nitrogen levels across a farm to monitor the health of crops (reprinted from~\cite{tokekar2016sensor} with permission).]{\includegraphics[width=0.40\columnwidth]{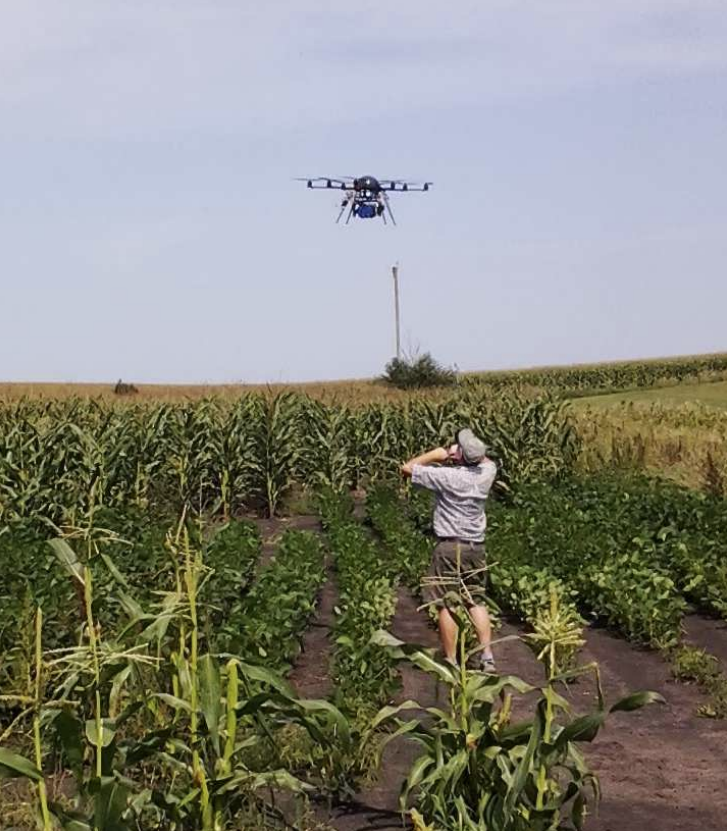}}
\caption{
Example monitoring tasks.
}
\label{fig:monitoring}
\end{figure}

The goal of monitoring is to observe and track an environment over time, as opposed to the learning task, which aims to acquire a globally consistent representation of the environment (refer to the example tasks in Figure~\ref{fig:monitoring}).

\subsection{Persistent monitoring}\label{subsec:monitoring-persistent}
\emph{Persistent monitoring} seeks a strategy to continuously observe environments that frequently change, with the task spanning an indefinite or prolonged period. This goal is hard to achieve in practice due to the limited budget (\eg, battery life and time); several methods are proposed to increase the length of monitoring time. 
\citet{ding2019heuristic} take mobile depots into account for recharging persistent robots and present a method for finding the minimum-cost tour.
\citet{plonski2017environment} address the issue of unmanned surface vehicles tracking invasive fish under unknown obstacles through energy-efficient exploration based on the solar map, allowing for energy harvesting.

We present several persistent monitoring methods in the literature.
\citet{smith2011persistent} propose a method for persistent monitoring using underwater vehicles. Their objectives are to minimize deviation from the planned path due to ocean currents and maximize the information value.
\citet{smith2012persistent} investigate persistent monitoring in changing environments and introduce a controller for multiple robots as a linear combination of a finite set of basis functions.
\citet{palacios2016distributed} develop a distributed coverage estimation and control scheme for persistent monitoring.

Multi-robot \emph{patrolling} is a task for multiple robots to regularly visit predefined regions of interest. 
\citet{iocchi2011multi} assess the performance of state-of-the-art patrolling algorithms in practical applications where the environment exhibits linear, cyclic, and mixed shapes.

The goal of \emph{boundary detection} is to monitor the boundaries of an unknown environment for area size estimation.
\citet{matveev2015robot} propose a sliding mode control method for monitoring unknown and time-varying environmental boundaries.

\subsection{Information gathering}\label{subsec:monitoring-information}

An information-gathering task allows active monitoring and exploration of an environment by minimizing information-theoretic statistics, such as entropy and mutual information. 
\citet{ma2018data} propose an information-theoretic dynamic data map that captures an environment's temporal properties using Gaussian processes.
\citet{corah2018distributed} introduce a distributed sequential greedy assignment for online exploration and mapping using multi-robot teams. Their algorithm approximates the sequential greedy algorithm~\citep{singh2009efficient} efficiently when dealing with many robots.
\citet{bourne2020decentralized} investigate information-theoretic control for multi-agent systems and propose an efficient method for computing mutual information using dynamic programming and multi-threading techniques.
We present more details regarding information gathering in Section~\ref{subsec:info}.

%%%%%%%%%%
\chapter{Decision-Theoretic Optimization to Solve Tasks}\label{chap:decision}
%This section presents various types of decision-making problems that may arise in environmental monitoring.
Given an environmental monitoring task, a robotic researcher should determine an appropriate representation of the environment, formulate a desirable objective function, and develop algorithms to achieve the task. We discussed various environment representation candidates in Section~\ref{sec:rep}. Most environmental monitoring tasks are, in fact, decision-making problems as a robot must make a number of local or global decisions. Decisions can be made offline before the process starts, or online as the robot obtains new information during the process. Researchers design algorithms to determine decision variables in a proposed objective function, in order to exactly or approximately maximize a task utility. 

In particular, we consider decision-making problems from a 5W1H perspective: (1) where to take measurements from; (2) which tasks to be assigned; (3) what samples to collect; (4) when to collect samples; (5) how to learn environment; and (6) who to communicate.
We introduce detailed decision-theoretic approaches in the literature based on this 5W1H perspective and present how these approaches have been employed in existing studies. 

\begin{sidewaysfigure}
\centering
\includegraphics[width=\textwidth]{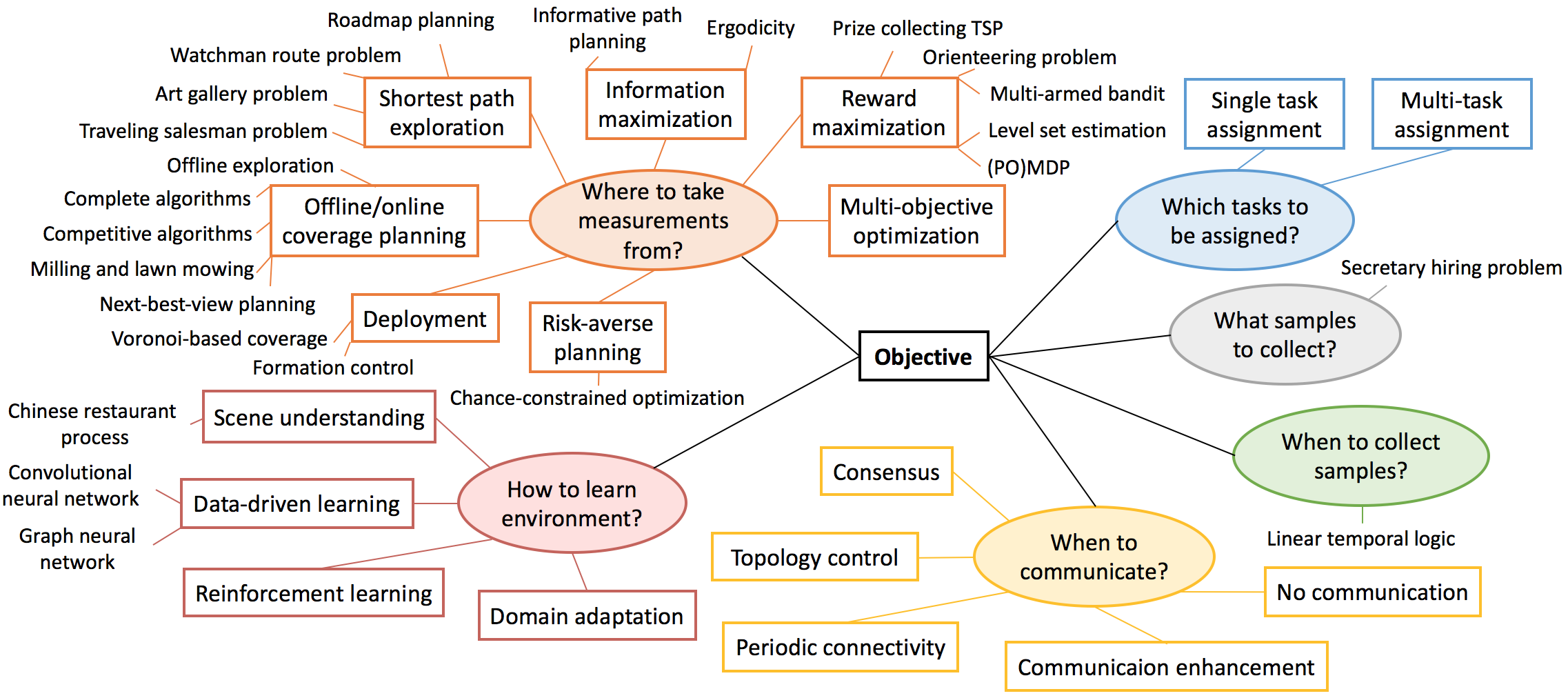}
\caption{Taxonomy of decision-theoretic approaches, categorizing a diverse range of methods and algorithms under overarching objectives. This taxonomy emphasizes the relationships between various tasks, decision-making methodologies, and their corresponding scientific objectives.}
\label{fig:diagram}
\end{sidewaysfigure}

\section{Where to Take Measurements from}~\label{sec:where}
To learn an underlying phenomenon of a given environment, robots must collect measurements (\ie, sampling) from the environment while adhering to time or energy constraints. The problem of where to take measurements from the environment can thus be considered as a path planning problem, associated with discrete sampling events along the robot path. 

Among many variants of path planning problems in the literature, we focus on the following ones: (1) shortest path exploration; (2) information gathering; (3) reward collecting; (4) online and offline coverage planning; and (5) deployment for multi-robot systems.

\subsection{Shortest path exploration}\label{subsec:shortest}
For environmental monitoring, one of the desired objectives is to find a minimum length path while visiting all sampling locations under certain constraints, such as collision avoidance and limited sensing capability. 
\NEW{
This problem can be generally formulated as:
\begin{equation*}
\begin{aligned}
\argmin_p \quad &L(p), \\
\textrm{s.t.} \quad &g_i(p)\ge0,\ \forall\ i, \\
&h_j(p)=0,\ \forall\ j,
\end{aligned}
\end{equation*}
where $p$ denotes a robot path, $L$ is a scalar function to quantify path length, the inequality constraints $g_i$ ensure collision avoidance and/or limited sensing capability, and the equality constraints $h_j$ guarantee that the robot visits all the sampling locations of interest. 
}
The solution of this optimization naturally results in a minimum time tour or a minimum energy tour as the objective is a function of a path length. 

When sampling locations are represented as a set of points, \emph{traveling salesman problem} (TSP)~\citep{applegate2006traveling} can be employed to find a shortest path that visits every sampling location exactly once for monitoring. As TSP is NP-hard~\citep{applegate2006traveling}, finding an optimal solution is challenging, but many approximate algorithms can find a reasonable solution. 
\emph{TSP with neighborhoods}~\citep{dumitrescu2003approximation,tokekar2016sensor,faigl2017autonomous} allows each sampling location to have a certain area instead of a point, assuming that values near a sampling location are similar. Visiting any point in an area implies that the corresponding sampling location is monitored; this implication relaxes the requirement of visiting a sampling location precisely. TSP with neighborhoods finds approximate solutions for the case when objective areas can be either overlapped or non-overlapped.
\emph{Generalized TSP}~\citep{noon1993efficient,yu2019algorithms} is another variant of TSP where the task is given by clusters of sampling locations. One method of obtaining a cluster of sampling locations is to sample continuous locations. The goal is to find a minimum length path that visits exactly one sampling location from each cluster. \citet{faigl2019fast} recently propose heuristics for approximate solutions to the generalized TSP with neighborhoods where the objective is to visit multiple target locations in 3D environments.

A major limitation of TSP-based data collection introduced above is that all sampling locations are assumed to be stationary and known. Some previous methods were proposed for moving nodes in a TSP graph~\citep{smith2009dynamic,hammar1999approximation,helvig2003moving} but had limited settings, \eg, all nodes moving with the same speed and direction. Most of them did not hold for an unknown and varying number of sampling locations. To address such challenges, stochastic variants of TSP have been proposed, such as stochastic radius of nodes for TSP with neighborhoods~\citep{kamousi2013euclidean}, stochastic node locations~\citep{citovsky2017tsp}, and stochastic occurrence of visiting a node~\citep{pavone2009stochastic}.

The above TSP-related research either assumes that the robot sensing ability is short-sighted or does not consider sensing distance. Some approaches still have a restricted assumption that the robot can observe any points in the environment as long as no obstacles exist between them, effectively giving it a $360\degree$ and unlimited sensing range, but exhibit desirable algorithmic guarantees in the spirit of computational geometry. We introduce recent approaches that alleviate these sensing range restrictions.

The \emph{art gallery problem}~\citep{o1987art} addresses the static sensor placement challenge. The objective is to identify the smallest set of sensing locations in a given environment, ensuring that every point within the environment is visible from at least one selected sensing location. 
\citet{arain2015efficient} introduce fan-shaped sensing regions using convex relaxation, allowing for overlapping field-of-view regions from multiple sensing locations. Once a set of sensing locations is determined, they employ a TSP solver to find a sequence of sites as a robot trajectory, commonly used in the art gallery problem for robotic applications.

In contrast to the art gallery problem that does not involve sensor mobility, its special variant, \emph{watchman route problem}~\citep{chin1986optimum}, is applicable to the mobile robotics domain. 
\citet{tokekar2015visibility} propose an optimal algorithm and a constant factor approximation to observe a set of target points using multi-robot systems equipped with an omnidirectional camera of unlimited sensing range. The same visibility restrictions are considered in \citet{zhang2019tree} to address the problem of maximizing the visibility of a robot in the presence of an adversarial target. Previous methods relax the visibility conditions by restricting a sensing distance to $d$: (1) \emph{$d$-watchman route
problem}~\citep{tan2003finding} where only the polygonal boundary of the environment is observed, and (2) \emph{$d$-sweeper route problem}~\citep{ntafos1992watchman} where a circular robot of radius $d$ sweeps the environment. A multi-robot version with $d$-visibility is studied in \citet{faigl2010approximate}.

Among motion planning algorithms, \emph{roadmap} methods developed for path planning (\eg, probabilistic roadmap method;~\cite{kavraki1996probabilistic}) have been adopted for environmental monitoring. A set of nodes is randomly generated from the configuration space of a robot to determine whether the robot is in the free space. Graph search techniques are used to find feasible paths from a start configuration to a goal configuration and choose the shortest path among the found paths.  
\citet{mishra2016battery} improve the scalability of the roadmap-based approach in handling battery constraints. In motion planning, various decomposition techniques besides roadmap planning have been proposed and adapted to environmental monitoring, such as boustrophedon cell decomposition~\citep{strimel2014coverage} and Morse decomposition~\citep{acar2002sensor}. For a more detailed survey on motion planning algorithms, see~\citet{galceran2013survey}.

\subsection{Information maximization}\label{subsec:info} 
Information gain is frequently used to determine where to take measurements in environmental monitoring. 
\NEW{The problem of maximizing information gain for a robot can be phrased as $\max F(p)$, where $p$ represents a path, and $F$ is a scalar function that quantifies the information collected along the path. This concept is known as \emph{informative path planning} (IPP).}

An exploration process has been interpreted as a task of reducing the uncertainty within the environment (\eg, uncertain knowledge about occupancy). The uncertainty is represented by fundamental quantities in information theory, such as entropy and mutual information. When an entropy value is assigned to each point in the environment, a higher value implies more uncertainty in the accessibility of the point~\citep{bhattacharya2014multi}. On the other hand, mutual information evaluates the (expected) information gain of unsensed locations with respect to a specific configuration of robots~\citep{bai2016information}. The goal of IPP is to generate paths for robots by actively sensing environmental phenomena, either by minimizing entropy or maximizing mutual information. 

\citet{krause2008near} propose a simple greedy algorithm that sequentially chooses sensing locations to maximize the mutual information, providing a formal guarantee of near-optimal performance by exploiting the \emph{submodularity}~\citep{nemhauser1978analysis}.
The notion of submodularity has been exploited often in this line of research. A set function $F:2^V\rightarrow \mathbb{R}$, where $V$ denotes a set of sampling locations, is submodular if it satisfies: $F(A\cup x)-F(A)\ge F(B\cup x)-F(B)$, where $A\subseteq B$ and $x\not\in A,\ B$. If a chosen utility function for an environmental monitoring task follows a submodular set function $F$ and satisfies \emph{monotonicity}, \citet{nemhauser1978analysis} shows that a simple sequential greedy algorithm can identify a near-optimal set of sampling locations, satisfying that $F(A_{\rm greedy})\ge (1-1/\epsilon) F(A_{\rm opt})$.

\citet{krause2008near} have also proposed several variations to study different aspects of the general submodular maximization problem. To overcome \emph{non-adaptability}, \citet{golovin2011adaptive} design a greedy algorithm for adaptive submodular maximization, allowing policy adjustments based on currently available information.
They also develop an exploration-exploitation framework for nonmyopic active learning of the Gaussian process. The theoretical bound depends on the difference between active learning and a priori design strategies~\citep{krause2007nonmyopic}. Exploration aims to reduce the uncertainty about model parameters, while exploitation involves identifying a near-optimal sampling policy.
In their subsequent work, \citet{krause2008robust} focus on minimizing the maximum posterior variance in Gaussian process regression for outbreak detection, demonstrating the \emph{robustness} of their algorithm and showcasing its efficient performance with approximation guarantees.
\citet{calinescu2011maximizing} investigate monotone submodular maximization under \emph{matroid constraints}.

Additional practical challenges have also been studied in the IPP problem. 
\citet{binney2013optimizing} consider time-varying fields and edge-based samples, allowing the robot to collect samples while moving from one node in the search graph to another.
\citet{atanasov2014information} design a non-myopic IPP algorithm that incorporates sensor dynamics, demonstrating suboptimality based on concavity.
\citet{saulnierinformation} propose an active exploration algorithm based on truncated signed distance fields. They optimize a sensor trajectory using a deterministic tree search to evaluate information gain and employ branch-and-bound pruning for an efficient search.

Similar to the motivation in adaptive submodular maximization in~\citet{golovin2011adaptive}, previous work has researched online IPP that modifies the current plan adaptively as new information is obtained. 
\citet{ma2018data} propose a planning and learning algorithm for persistent monitoring tasks. Their method plans a trajectory that maximizes mutual information while learning a Gaussian process-based datamap to model a spatiotemporal field and tuning hyperparameters online.
\citet{popovic2020informative} study the monitoring objective of mapping discrete and continuous variables on the terrain within the IPP framework. They also concentrate on strategies to adapt plans online for different scenarios. 
\citet{schmid2020efficient} introduce an optimal motion planning-driven IPP algorithm to address the suboptimality often associated with sampling-based methods.

Multi-robot systems have also been investigated for more efficient IPP solutions.
\citet{cao2013multi} develop an active sensing IPP algorithm for multiple robots based on entropy and mutual information criteria. They prove guarantees on the trade-offs between active sensing performance and time efficiency.
\citet{corah2018distributed} extend the previous sequential greedy algorithm that works only for a single robot to multiple robots and provide a near-optimality guarantee for the multi-robot version. 
\citet{nguyen2016information} propose a resource-constrained adaptive sampling strategy for a team of mobile robots. Based on conditional entropy, their method comes with a theoretical guarantee regarding both upper and lower bounds for the solution.
\citet{low2008adaptive} present an adaptive model-based exploration method that plans non-myopic multi-robot paths by minimizing the expected sum of posterior variances over all locations.
\NEW{Minimum makespan, minimum number of measurement locations, and minimum total time multi-robot path planning using Gaussian process regression have been studied in~\citet{suryan2020learning}.}
Additional challenges have also been considered within the scope of multi-robot IPP, such as uncertainty of a hidden state over unknown horizon~\citep{kantaros2021sampling}, a phenomenon of interest not being able to be observed directly~\citep{malencia2022adaptive}, adversarial environments~\citep{schlotfeldt2021resilient}, unreliable communication among robots~\citep{woosley2020multi}, and limited communication ranges among robots and control uncertainty~\citep{dutta2021opportunistic}.

The multi-objective problem in information maximization has been studied by \citet{gautam2017science} that minimizes the path length while maximizing the information gain for budgeted exploration. They use evolutionary algorithms to solve this Pareto optimization.
\citet{meera2019obstacle} propose the problem of maximizing the information gain and minimizing the collision cost.

\emph{Ergodic theory} regards time-averaged behavior along the trajectory of a dynamical system with respect to the space of all possible states of the system~\citep{miller2015ergodic}. Ergodicity can be used as another information metric for environmental monitoring where a robot distributes the allotted time to regions proportional to the expected information gain in those regions. Ergodicity can also address both exploration and exploitation for coverage using the expected information density. Ergodicity, in contrast to other information metrics, naturally encourages spending more time exploring nearby regions with high expected information rather than solely visiting the region with the highest information. Previous work has shown the effectiveness of ergodicity in terms of active sensing, including trajectory optimization of ergodic exploration for nonlinear and deterministic control systems~\citep{miller2015ergodic}, receding-horizon ergodic exploration for both coverage and target localization~\citep{mavrommati2017real}, and ergodic exploration for multi-robot systems~\citep{abraham2018decentralized}. 

\subsection{Reward maximization}\label{subsec:reward}
When collecting a sample is associated with a reward, environmental monitoring can be addressed as a reward maximization problem. The goal is to find a trajectory that maximizes the rewards from collected samples while satisfying constraints, such as a limited energy budget, if it exists. 
\NEW{
The general form of reward maximization is represented as follows:
\begin{equation*}
\begin{aligned}
\argmax_p \quad &R(p), \\
\textrm{s.t.} \quad &L(p)\le B,
\end{aligned}
\end{equation*}
where $p$ represents a robot path, $R$ is a scalar function that measures the rewards, $L$ is a scalar function that quantifies the path length, and $B$ denotes a given budget.
}
Additional objectives studied in reward maximization include maximizing a robot lifetime~\citep{jadaliha2013environmental} and reducing resource usage~\citep{parker2013estimation}.

Reward maximization has been studied in computational geometry. \emph{Prize collecting} traveling salesman problem~\citep{balas1989prize} finds a subset of cities to visit such that travel costs and net penalties--getting a prize when visiting a city and a penalty when failing to visit a city--are minimized.

Reward maximization can be interpreted as cost minimization from the control theory perspective. An optimization-based approach has been studied to minimize certain cost functions for optimal data collection.
\citet{leonard2007collective} present a data collection scheme to compute optimal multi-robot paths by minimizing the error in a model estimate of the sampled field.

Another variant is the \emph{orienteering problem}, where the goal is to maximize the reward collected until a given budget (\eg, time or fuel) is exhausted.
\citet{chekuri2005recursive} propose a first recursive greedy algorithm for the orienteering problem, where the objective function is a submodular set function. They prove that their algorithm runs in quasi-polynomial time and provides an approximation guarantee of $\mathcal{O}(\log \rm{OPT})$, where $\rm{OPT}$ denotes the optimal value for a given problem.
Various methods in the robotics community are built upon the results of this recursive greedy algorithm to develop \emph{submodular orienteering} algorithms, such as \citet{binney2013optimizing}.
The previous work mostly consider the reward function of a sampling location to be independent of rewards in other locations.
\citet{yu2016correlated} propose a correlated orienteering problem where a quadratic utility function is considered to capture a spatial correlation.
Other practical challenges have been addressed in the orienteering problem, such as orienteering with neighborhoods~\citep{faigl2016self}, service time dependant rewards~\citep{mansfield2021multi}, uncertain rewards~\citep{liu2021team}, unknown rewards~\citep{wilde2021learning}, and uncertain environment model~\citep{peltzer2022fig}.

Another problem, closely related to the orienteering problem, is the \emph{optimal tourist problem}. Compared to the orienteering problem, the optimal tourist problem treats the reward function as a function of the time spent at the sampling location. \citet{yu2015anytime} investigate the optimal tourist problem based on two complementary problems. The first problem solves the same objective as in the orienteering problem, and the second problem minimizes the time for gathering a predetermined reward amount.

\emph{Multi-armed bandit} (MAB) regards the problem of finding an arm associated with an unknown black-box function out of a fixed number of arms to maximize the expected gain (or reward) when a finite number of arm evaluations is allowed. MAB naturally addresses the exploration versus exploitation dilemma. To design algorithms with useful theoretical bounds in MAB, the concept of \emph{regret} is created to measure the expected gap of the collected rewards between the optimal strategy and a designed algorithm. The \emph{upper confidence bound} (UCB) algorithm is a well-known algorithm that gives a theoretical upper bound on regret, owing to the principle of optimism in the face of uncertainty. One can choose where to evaluate the black-box function next according to UCB. 
\citet{reverdy2014modeling} propose MAB with transition costs and on graphs appropriate for robotics applications. \citet{landgren2016distributed} examine distributed MAB for cooperative decision-making based on the consensus algorithm. However, in this setting, the black-box function of one arm is independent of that of neighboring arms.

Among various approaches to solving MAB, \emph{Bayesian optimization} is often used for robotics environmental monitoring as it considers spatial correlation between arms. Bayesian optimization adopts a sequential sensor placement strategy for global optimization of unknown black-box functions, used for Bayesian experimental design and active learning. Bayesian optimization version of the UCB algorithm is proposed by \citet{Srinivas:2010:GPO:3104322.3104451}. They use the Gaussian process (GP) to model the correlation of the environment, called GP-UCB. The acquisition function (\ie, surrogate objective function) to determine which location to sample next follows this functional form: $\mu(x) + \kappa\sigma(x)$, where $\mu(x)$ and $\sigma(x)$ are the mean and the standard deviation at location $x$, respectively. The weight parameter $\kappa$ balances the trade-off between exploration and exploitation. The next sampling location $x$ can be determined by maximizing the acquisition function.

\citet{marchant2012bayesian} also incorporate the GP in Bayesian optimization for robotics environmental monitoring. They design a new acquisition function, called distance-based UCB, to reduce the total travel distance. The adaptive sampling strategy proposed by~\citet{tan2018gaussian} is based on cross-entropy trajectory optimization, where the objective is the same as GP-UCB. 
When unmanned aerial vehicles are considered, the \emph{multi-fidelity GP-UCB}~\citep{kandasamy2019multi} explores different fidelity levels of sample values observed at various altitudes of the robot~\citep{wei2020expedited,sung2021environmental}.

In some cases, the monitoring task would not necessarily be aimed to find maximum value points in an environment, but to find regions where values of points in regions are above a predefined threshold value. The estimation of those regions is referred to \emph{level set estimation} (LSE).
\citet{gotovos2013active} employ GP to model the underlying function and form LSE as a sequential decision problem, which is inherently similar to GP-UCB~\citep{Srinivas:2010:GPO:3104322.3104451}. They also present two extensions to LSE where the first extension considers the threshold level implicitly defined as a function of the maximum value, and the second one gathers a batch of samples. 
\citet{hitz2014fully} explores the path planning aspect for the LSE algorithm to monitor temporal and spatial dynamics of harmful cyanobacterial blooms in lakes. They leverage dynamic programming to develop a receding horizon path planner for LSE. 

In sequential decision-making, \emph{Markov decision process} (MDP) computes an optimal policy for a robot when the transition model and reward model are known. If the state of the target function is not fully observable, a belief can be used to represent the state, \ie, \emph{partially observable} MDP (POMDP). 
\citet{marchant2014sequential} formulate sequential Bayesian optimization as POMDP. They apply \emph{Monte-Carlo tree search} (MCTS) and UCB to trees to handle continuous state and observation spaces for spatial-temporal monitoring.
\citet{morere2017sequential} redefine the reward function as a trade-off between exploration and exploitation of gradients in the form of Bayesian optimization POMDP. They employ GP and MCTS to allow continuous non-myopic planning. 
Because MCTS is limited to discrete actions, in their subsequent work~\citep{morere2018continuous}, they propose dynamically sampling promising actions based on MCTS to achieve continuous action spaces. This approach is incorporated by kernel-based trajectory generation from the theory of reproducing kernel Hilbert spaces, yielding the properties of smoothness and differentiability.
\citet{flaspohler2019information} propose a POMDP algorithm to localize and collect globally maximal samples from unknown and partially observable continuous environments.
\citet{lauri2020multi2} study multi-agent active information gathering by policy graph improvement in a POMDP setting.

\subsection{Offline/online coverage planning}\label{subsec:offline} 
Robotic coverage is often used to observe all points at least once in environmental monitoring. In most cases, a finite field-of-view (FOV) of the sensor and deterministic sample values are considered with a focus on designing a robot trajectory that covers the entire environment in minimum time. 
\NEW{
That is, $\argmin_p L(p)\ \textrm{s.t.}\ C(p)=\texttt{True}$, where $p$ represents a path, $L$ is a scalar function that measures the path length, and $C$ is a Boolean function that returns \texttt{True} if the environment is fully covered by the trajectory of the robot FOV, or \texttt{False} otherwise.
}
Previous work has analyzed the algorithmic aspect of their proposed algorithms, such as \emph{completeness} and \emph{competitiveness}. Coverage planning can be categorized depending on whether the information (\eg, size and shape) of an environment is a priori (\ie, \emph{offline exploration}) or obtained on the fly (\ie, \emph{online exploration}).

In offline exploration, the size and shape of the environment as well as the size of the FOV of the sensor are given. If a designed algorithm is an approximation algorithm, \emph{approximation ratio} is usually investigated, \ie, the performance ratio between the approximation algorithm and optimal solution. When the sensor has a symmetrical 2D FOV, such as circular FOV, \emph{Boustrophedon cellular decomposition}~\citep{choset2000coverage}, also known as lawn mowing~\citep{bormann2018indoor}, finds a complete coverage path.
\citet{strimel2014coverage} propose complete battery-constrained sweep planning for a single robot, which is an extension to a general boustrophedon decomposition. This method requires an assumption that a single battery and recharging station exist to allow a robot to completely cover the environment. 
\citet{wei2018coverage} revisit a similar problem where the robot is allowed to charge at the recharging station, but their objective is to minimize the number of paths (\ie, the new path starts after recharging) and then to minimize the path length as the second objective. Their approach is based on the traveling salesman problem over the grid map. They show the approximation ratios of $4$ and $8$ for minimizing the number of paths and the path length, respectively, for contour-connected environments. For arbitrary environments, they prove $4(2r+4)$ and $8(2r+4)$--approximation ratios, where $r$ is the number of reflex nodes.
\citet{hameed2014intelligent} develops an energy-efficient coverage algorithm in 3D that considers terrain inclinations and determines the optimal driving direction by minimizing the total energy consumption using a genetic algorithm.

In the case of multi-robot exploration, optimizing the number of robots or the amount of energy required is proven to be NP-hard~\citep{fraigniaud2006collective}.
\citet{mishra2016battery} study battery-constrained coverage for multiple robots based on the roadmap approach. They consider the heterogeneous behavior among robots including worker and helper robots. Their algorithm computes how long robots can explore before recharging without having to set the battery threshold explicitly.
\citet{karapetyan2018multi} propose heuristic algorithms for multi-Dubins vehicles to evenly distribute workloads among robots.
\citet{agarwal2020line} design a heuristic-based approximation algorithm for multiple robots for line coverage, where the robots must visit all points of linear environment features such as road networks and power lines.
The continuous 2D polygonal environment has been studied by \citet{farias2020planar}, where they propose multi-robot navigation from source to sink by computing the continuous maximum flow. 

For online exploration, no prior information about an environment is available, and therefore, a robot needs to adapt its plan as new information is received on the fly. Some algorithms are incomplete, \ie, not guaranteeing complete coverage.
\citet{sim2009autonomous} propose vision-based exploration strategies in combination with SLAM and utilize a hybrid map for both localization and safe navigation.
\citet{cesare2015multi} study frontier-based exploration for multiple UAVs under limited communication and battery constraints. They introduce heterogeneous behavior for efficient planning, allowing robots to switch between four states (\ie, explore, meet, sacrifice, and relay).
\citet{hassan2019ppcpp} consider a scenario where unexpected changes occur, such as the sudden appearance of stationary or dynamic obstacles in an environment. 

When a given problem is NP-hard in online setting, competitiveness is a useful algorithmic guarantee that measures a theoretical gap from the unknown optimal solution. The competitive ratio is the worst-case ratio between the performances of a proposed online algorithm and the optimal offline algorithm. Previous work proposed complete algorithms but without a competitive guarantee.
\citet{bender2002power} address the problem of exploring and mapping an unknown environment when node labels are not given. They introduce the concept of a \emph{pebble} to identify a node label. While their approach is designed for a single robot, they prove that only one pebble is sufficient if an upper bound on the number of nodes (denoted by $n$) is known, or $\Theta(\log\log n)$ pebbles are both necessary and sufficient otherwise. 
\citet{das2007map} propose a strategy to construct a labeled map by initially dispersing robots in two stages: first, robots explore and mark nodes, and second, they share information to combine the partial maps. Their analysis relies on the assumption that the graph size is co-prime with the number of robots, resulting in a deterministic solution regardless of the graph's structure.
\citet{jensen2019effects} investigate different communication models, ranging from global communication to no direct communication, and evaluate the proposed algorithms empirically.
The problem proposed by~\citet{osswald2016speeding} does not involve exploring a completely unknown environment. Instead, they work with a topo-metric map that provides background information given by the user. They propose a faster exploration strategy based on TSP by leveraging this information.
\citet{lee2016structured} design distributed exploration algorithms for multiple robots based on triangulating an unknown environment. They address two related problems: minimizing the number of robots to cover the entire environment when the number of available robots is not bounded a priori, and maximizing the covered area when the number of available robots is fixed. They show that the competitive ratio for the former problem is 3, while the latter does not have a competitive ratio. They also provide performance guarantees for the coverage of the environment by exploiting the dual graph of the triangulation with respect to the original geometric space.
\citet{umari2017autonomous} study unknown environment exploration based on multiple rapidly-exploring randomized trees. They construct local and global trees for efficient frontier detection.

We present a list of competitive algorithms for multi-robot online exploration.
Usually, a tree structure, \ie, tree exploration, is used to analyze the competitiveness. The competitive ratio differs depending on the definition of cost (\eg, the maximal time or the maximal energy taken to explore the environment) and the sensing model (\eg, one can observe up to (1) incident edges from the current node and (2) neighboring nodes).
For communication among robots, two models are typically assumed: global communication, allowing robots to exchange information at any time, and local communication, enabling information exchange when robots are on the same node simultaneously~\citep{das2015collaborative}. Another popular communication model is to introduce bookkeeping devices that can write and read information on visited nodes. Let $n$ be the number of robots.
\citet{fraigniaud2006collective} call this \emph{write-read communication} and propose a $\mathcal{O}(\frac{n}{\log{n}})$--competitive algorithm that minimizes the exploration time with the sensing model (1). They also prove the lower bound of $2-1/n$. 
\citet{brass2011multirobot} also use the same sensing model and improve the competitive ratio to $2e/n+\mathcal{O}((n+r)^{n-1})$, where $e$ and $r$ denote the number of edges and the radius, respectively.
\citet{dynia2007robots} improve the lower bound to $\Omega(\frac{\log{n}}{\log{\log{n}}})$. They also propose a $4-2/n$--competitive algorithm when the cost is the maximal energy of a robot.
\citet{higashikawa2014online} develop a $\frac{n+\floor{\log{n}}}{1+\floor{\log{n}}}$--competitive algorithm of multi-robot exploration for a static plume when the time cost is considered under the local communication. This competitive ratio is extended by \citet{sung2019competitive} for a translating plume to $\frac{2(s_r+s_p)(n+\floor{\log{n}})}{(s_r-s_p)(1+\floor{\log{n}})}$, where $s_r$ and $s_p$ are the speed of the robot and that of the plume, and to $\frac{2(s_r+s_p)(18n+\floor{\log{n}})}{(s_r-s_p)(1+\floor{\log{n}})}$ if the plume shape is not restricted to grid cells but an arbitrary shape.
\citet{preshant2016geometric} study the environment that is an orthogonal polygon, proposing the competitive ratio of $\frac{2(\sqrt{2}n+\log{n})}{1+\log{n}}$.
\citet{megow2012online} analyze the competitive ratio for any general graphs with arbitrary weights using the DFS. They also study the cases of general graphs with bounded genus.
\citet{das2015collaborative} address the problem of minimizing the team size, given limited energy, that can explore an unknown environment. Denoting the amount of energy as well as the maximum possible tree height by $B$, they show a $\mathcal{O}(\log{B})$--competitive algorithm.
\citet{mahadev2017mapping} develop a $\mathcal{O}(\log{D}/\log\log{D})$--competitive algorithm in $1$ dimension, given an unknown distance $D$, for MR imaging while considering two different and independent costs for moving and for measuring.
\citet{shnaps2016online} study battery-constrained sweep planning. The robot starts from the recharging station, returns when the battery level is low, and continues the coverage task after recharging. They propose $(L/S)$--competitive algorithm for a mobile robot, where $S$ is the size of the robot and $L$ is the path length that the robot can move with a full battery charge. 
They also analyze the competitive ratio for online tethered coverage where a robot of size $D$ is tethered at a fixed point by a cable of length $L$~\citep{shnaps2014online}. They prove upper and lower bounds on the competitive ratio.
For the sake of the complete survey, we summarize the competitiveness of various problems and algorithms in Table~\ref{tab:online}.

\begin{landscape}
\centering
\begin{longtable}{|c|l|c|}
\hline
\textbf{Ref.} & \multicolumn{1}{c|}{\textbf{Problem}} & \textbf{CR}\\ \hline
\multirow{4}{*}{\small{\citet{lee2016structured}}} & Minimizing the number of robots for covering all of & \multirow{2}{*}{$3$} \\
& the region & \\
\cline{2-3}
& Maximizing the covered area with a fixed number of & \multirow{ 2}{*}{Not provided} \\
& robots & \\
\hline
\multirow{3}{*}{\small{\citet{plonski2017environment}}} & Navigation around an unknown obstacle  & $\mathcal{O}(1)$ \\ \cline{2-3}
& Energy-efficient exploration of the solar map of the & \multirow{ 2}{*}{$\mathcal{O}(\log{c})$} \\
& environment & \\
\hline
\multirow{ 2}{*}{\small{\citet{fraigniaud2006collective}}} & \multirow{ 2}{*}{Minimizing the time of multi-robot tree exploration}  & $\mathcal{O}(\frac{n}{\log{n}})$ \\
\cline{3-3}
& & LB: $2-1/n$ \\
\hline
\multirow{ 2}{*}{\small{\citet{brass2011multirobot}}} & \multirow{ 2}{*}{Minimizing the time of multi-robot tree exploration}  & $2e/n+\mathcal{O}((n+r)^{n-1})$ \\
\cline{3-3}
& & LB: $\max(2e/n,2r)$ \\
\hline
\multirow{ 2}{*}{\small{\citet{dynia2007robots}}} & Minimizing the time of multi-robot tree exploration  & $\Omega(\frac{\log{n}}{\log{\log{n}}})$ \\ \cline{2-3}
& Minimizing the energy of multi-robot tree exploration & UB: $4-2/n$ \\
\hline
\multirow{ 2}{*}{\small{\citet{higashikawa2014online}}} & \multirow{ 2}{*}{Minimizing the time of multi-robot tree exploration}  & UB: $\frac{n+\floor{\log{n}}}{1+\floor{\log{n}}}$ \\
\cline{3-3}
& & LB: $\ceil{\frac{n}{1+\floor{\log{n}}}}-\epsilon$ \\
\hline
\multirow{ 4}{*}{\small{\citet{sung2019competitive}}} & Minimizing the time of multi-robot tree exploration & UB: \\
& of a translating plume & $\frac{2(s_r+s_p)(n+\floor{\log{n}})}{(s_r-s_p)(1+\floor{\log{n}})}$ \\
\cline{2-3}
& Minimizing the time of multi-robot tree exploration of & UB:\\
& a translating arbitrarily-shaped plume & $\frac{2(s_r+s_p)(18n+\floor{\log{n}})}{(s_r-s_p)(1+\floor{\log{n}})}$  \\
\hline
\multirow{ 2}{*}{\small{\citet{preshant2016geometric}}} & Maximizing the coverage area where the envionment & UB: $\frac{2(\sqrt{2}n+\log{n})}{1+\log{n}}$ \\
\cline{3-3}
& is an orthogonal polygon & \\
\hline
\multirow{4}{*}{\small{\citet{megow2012online}}} & Minimizing the time of multi-robot exploration of & \multirow{2}{*}{UB: $2(2+\epsilon)(1+2/\epsilon)$} \\ 
& general graphs with arbitrary weights & \\
\cline{2-3}
& Minimizing the time of multi-robot exploration of & UB: \\
& general graphs with bounded genus & $2(2+\epsilon)(1+2/\epsilon)(1+2g)$ \\
\hline
\multirow{ 2}{*}{\small{\citet{das2015collaborative}}} & Minimizing the team size of multi-robot tree & $\mathcal{O}(\log{B})$ \\
\cline{3-3}
& exploration given a limited energy & $\Omega(\log{B})$ \\
\hline
\multirow{ 2}{*}{\small{\citet{mahadev2017mapping}}} & Minimizing the moving and measuring costs of & \multirow{ 2}{*}{$\mathcal{O}(\log{D}/\log\log{D})$} \\
& multi-robot exploration in 1 dimension & \\
\hline
\multirow{ 2}{*}{\small{\citet{shnaps2016online}}} & Minimizing the time of a single-robot battery- & UB: $L/S$ \\
\cline{3-3}
& constrained coverage when recharging is allowed & LB: $\log{(L/4S)}$ \\
\hline
\multirow{2}{*}{\small{\citet{shnaps2014online}}} & Maximizing the coverage area using a robot of size $D$ & UB: $2(L/D)$ \\
\cline{3-3}
& when tethered at a fixed point by a cable of length $L$ &LB: $\log{(L/D)}$ \\
\hline
\multirow{ 2}{*}{\small{\citet{gul2022online}}} & Watchman route problem where there is a convex & \multirow{2}{*}{UB: $89.83$} \\
& polygonal obstacle & \\
\hline
\small{\citet{icking2005exploring}} & Online milling problem & UB: $\frac{4}{3}$ \\
\hline
\small{\citet{kolenderska2009improved}} & Online milling problem & UB: $\frac{5}{4}$ \\
\hline
\caption{Online exploration algorithms with the proposed competitive ratio (CR). $n$ is the number of robots; $c$ is the number of critical points to observe; $e$ is the number of edges; $r$ is the radius; $s_r$ and $s_p$ are the speed of the robot and that of the plume; $g$ implies that  graphs of genus at most $g$; $B$ denotes the amount of energy as well as the maximum possible tree height; $D$ is an unknown distance; $S$ is the size of the robot; $L$ is the path length that the robot can move with a full battery charge; and $\epsilon$ is a nonnegative value. UB and LB stand for the upper bound and the lower bound, respectively.}
\end{longtable}
\label{tab:online}
\end{landscape}

Another perspective to approach the coverage problem is whether the robot is constrained to move only within the boundary of the environment, or allowed to move outside the boundary. The former is \emph{milling}, and the latter is \emph{lawn mowing}. 
In milling,
\citet{arkin2000approximation} present $2.5$--approximation algorithm for offline milling with a single robot.
\citet{arya2001approximation} develop an approximation algorithm for offline milling for multiple robots.
\citet{icking2000exploring} propose a strategy that yields a competitive tour 
%of length $S$, such that $S\le C+\frac{1}{2}E+H-3$, where $C$, $E$ and $H$ denote the number of cells, that of edges and that of obstacles, respectively, 
for online milling of an polygon which may contain holes with a single robot.
\citet{icking2005exploring} present a $\frac{4}{3}$--competitive algorithm for online milling of polygons without holes.
\citet{kolenderska2009improved} improve upon this to give  an online milling algorithm with a competitive ratio of $\frac{5}{4}$.
In lawn mowing,
\citet{arkin2000approximation} present $(3+\epsilon)$--approximation algorithm for offline lawn mowing with a single robot.
\citet{reggente20103d} build
three-dimensional gas distribution maps from gas sensor and wind measurements obtained with a mobile robot.

\emph{Next-best-view} (NBV) planning is online planning to reconstruct a 3D environment, without assuming a prior model of the environment~\citep{bircher2018receding}, which has been extensively studied for 3D object reconstruction~\citep{vasquez2017view}. Although NBV planning is not precisely coverage planning, it often covers the environment as much as possible in pursuing better 3D reconstruction. Every time a robot receives a new scan, the robot evaluates a set of candidate viewpoints based on the sensor specification and the utility of each viewpoint. Once one of the viewpoints is chosen, the robot navigates to the designated viewpoint to obtain a new scan which will be incrementally registered to the previous scans. 
\citet{vidal2018optimized} propose quadtree-based map representation for efficient NBV planning. Subsequently, they develop an NBV planning algorithm for exploration of complex underwater environments~\citep{palomeras2019autonomous}.
\citet{meng2017two} design a two-stage NBV planning algorithm based on heuristic information gain in an unknown 3D environment. The two-stage algorithm consists of frontier-based boundary coverage and a pathfinder using a TSP solver. Their algorithm aims at both the exploration and reconstruction of 3-D unknown environments.
Most NBV planning copes with a single-object scene; \citet{heng2015efficient} propose an efficient planning algorithm achieving two objectives of exploration and coverage for multi-object scenes.
\citet{best2018online} address multi-robot path planning to observe a set of continuous viewpoint regions maximally. Their polynomial-time algorithm jointly finds the nodes and their sequence and allocates them to robots. 
\citet{lauri2020multi} formulate multi-sensor NBV planning as matroid-constrained submodular maximization. 
\citet{dharmadhikari2021hypergame} form a game between the environment and the robot such that the environment provides unknown future challenges, and the robot finds the best policy (\ie, exploration and visual search) without knowing which challenge it will face.

\NEW{The distinction between offline and online contexts is crucial in planning algorithms that utilize Gaussian process regression. This distinction is based on the availability of the Gaussian process's covariance function. Offline planning is suitable when the covariance function is known a priori, enabling exploration and planning optimization using the locations of future measurements. In contrast, online planning becomes crucial when the covariance function is unknown. The covariance function can be determined from actual sampling while adaptively computing a path. For further details on Gaussian process-based algorithms for environmental monitoring, refer to Section~\ref{sec:rep}.}

\subsection{Deployment}\label{subsec:deploy}
To determine how to cooperatively distribute measurements for multiple robots, \emph{deployment} algorithms have been developed to efficiently generate trajectories of robots. Among various deployment methods, \emph{Voronoi-based coverage} has drawn particular attention as it exhibits desirable behaviors for multiple robots such that algorithms are adaptive, distributed, asynchronous, and provably correct. Voronoi-based coverage is built upon Voronoi diagram where the environment is equidistantly partitioned into closest pairs of robots. 
Voronoi-based coverage is firstly proposed by \citet{cortes2004coverage}, called \emph{Lloyd’s algorithm}, where the goal is to find an optimal sensor allocation of the environment of interest. 
\NEW{Their method assumes a known probability density function that represents the likelihood of a specific event occurring within a convex polytope.}
Following the Voronoi diagram principle, this allocation can be achieved by partitioning the domain such that each robot is assigned to one partition.
\citet{bhattacharya2014multi} generalize the discrete-time Lloyd’s algorithm proposed by \citet{cortes2004coverage} to handle a non-convex environment in non-Euclidean space due to obstacles. They prove the stability of their control law  for coverage on Riemannian manifolds with boundaries.
\citet{elwin2019distributed} propose a distributed algorithm based on a finite element method that accounts for the spatial correlation between measurements and estimation locations. Their main focus is to reduce the memory consumption and communication requirements of robots by incorporating the spatial correlation.
However, optimally learning and covering a spatial field simultaneously remains an open problem, as addressed by \citet{santos2021multi}.
Additional challenges in Voronoi-based coverage have also been addressed, including
connectivity constraints in communication among robots~\citep{luo2019voronoi} and
energy constraints~\citep{jensen2020near}.

\NEW{Unlike Lloyd's algorithm that assumes a known distribution density function, \citet{le2012adaptive} consider that the event location distribution is a priori unknown and could only be progressively inferred from the observation of the actual event occurrences using distributed stochastic gradient algorithms.
\citet{zuo2017efficient} address the unknown density function in coverage control by developing an adaptive spatial estimation algorithm, integrating a consensus mechanism for efficiency, and proposing a distributed strategy to optimally position sensors. 
\citet{schwager2009decentralized} introduce a decentralized, adaptive control law for mobile robots that use sensor data and a consensus algorithm, validated through Lyapunov-type proofs and simulations.
\citet{carron2015multi} present a method where agents simultaneously estimate and optimally cover a region using centroidal Voronoi partitions within a Bayesian regression framework, without a previously known sensory function. 
Similarly, \citet{benevento2020multi} employ Bayesian Optimization with Gaussian Processes to simultaneously estimate and optimize coverage of an initially unknown spatial scalar field.
}
A detailed survey can be seen in~\citet{paley2020mobile}.

% In control theory, \emph{formation control}~\citep{oh2015survey} guides a team of robots to satisfy constraints in their states while allowing the group to move as a whole.
% Wu and Zhang~\citep{wu2011cooperative} proposed cooperative multi-robot exploration of level surfaces of unknown 3D scalar field based on formation control. The formation formed by robots tracks lines of curvature on a desired level surface in the field. They proved necessary and sufficient conditions for reliable estimates and a minimum number of robots required. 

\subsection{Risk-averse planning}\label{subsec:risk}
When monitoring hazardous environments, safe planning for robots is desirable to avoid risky situations, such as collisions with obstacles, while still achieving the task goal. 
\NEW{
This problem can be framed as a constraint feasibility problem with the objective of finding a feasible path from start to goal while adhering to specific risk-related constraints to ensure safety. Depending on how these constraints are interpreted, various mathematical optimization approaches can be developed.
}

Although worst-case scenarios have been analyzed in robust optimization and formal verification, they often results in a conservative plan. In contrast, various risk measures have been proposed, such as \emph{value at risk}~\citep{jorion2007value} and \emph{conditional value-at-risk}~\citep{rockafellar2000optimization}, for stochastic risk management. Here, we present existing work specific to environmental monitoring.

\citet{hernandez2019online} propose a risk objective function that incorporates both the length and safety of the path in underwater vehicle applications.
\citet{ono2014safe} investigate safe path planning for marine robots. To avoid capsizing and bow-diving, they propose a method of classifying safe and unsafe velocities. They then adopt model predictive control for planning a path.
\citet{papachristos2019localization} research belief space planning for aerial robotic uncertainty-aware path planning. Their algorithm ensures mapping consistency while exploring an unknown environment.
\citet{hollinger2016learning} propose a risk-aware Gaussian process to learn the uncertainty in ocean current prediction for safe navigation.
\citet{somers2016human} develop a risk-averse marine data collection method given by human operator's preferences but refined by probabilistic planning with nonparametric uncertainty modeling.
A more detailed survey can be found in~\citet{zhou2021multi}.

\emph{Chance-constrained optimization} is one of the optimization techniques where some constraints dictate the probability of having risky events being bounded.  \citet{ono2015chance} propose dynamic programming-based chance-constrained optimization that handles joint chance constraints caused by uncertain terrain conditions for space exploration.

Previous work has employed \emph{linear temporal logic} to consider the notion of safety encoded as user-defined specifications.
\citet{barbosa2019guiding} adopt signal temporal logic, extended from linear temporal logic to handle real-time and real-valued constraints, to explicitly model maximum and minimum allowable distances from obstacles.

\section{Which Tasks to be Assigned}~\label{sec:which}
In large-scale environmental monitoring, the size of the environment is significantly larger than that of the FOV of a sensor mounted on a robot, making the robot spend an enormous amount of time gathering useful information. Therefore, multi-robot systems would expedite task completion by distributing sensing burdens cooperatively. The deployment of multiple robots, furthermore, yields beneficial features over single robot deployment, such as resiliency against failures, reliability for complex tasks, and simplicity in design~\citep{khamis2015multi}.

%Give some references deploying multiple robots in the actual field for environmental monitoring.

In the multi-robot literature, discrete targets and disjoint regions are often called tasks. The critical challenge to multi-robot coordination from the planning perspective is the problem of assigning a set of given tasks to robots. Tasks must be allocated among robots to minimize time or energy spent by all robots and achieve desirable global task performance expressed in utility.

As task allocation has a vast amount of literature, we redirect interested readers to the survey papers~\citep{gerkey2004formal,korsah2013comprehensive,khamis2015multi}. In this survey, we highlight some of the recent work related to environmental monitoring. In particular, we present two constraints mainly studied in task allocation: each robot can be assigned to at most a single task at a time (\emph{single task assignment}) or multiple tasks (\emph{multi-task assignment}).

\subsection{Single task assignment}\label{subsec:single_task}
In the case of single-task robot and single-robot task, both robot and task are assigned to at most a single counterpart. This task can be formally formulated, shown in~\citep{gerkey2004formal}, as:
\begin{equation}~\label{eqn:single-task}
\begin{aligned}
\max \quad &\sum_{i=1}^M\sum_{j=1}^N\alpha_{ij}w_jU_{ij},\\
\textrm{s.t.} \quad & \sum_{i=1}^M\alpha_{ij}=1,\ \forall\ j\in\{1,..,N\}, \\
  &\sum_{j=1}^N\alpha_{ij}=1,\ \forall\ i\in\{1,..,M\}, \\
\end{aligned}
\end{equation}
where $i$ is the robot index, $j$ is the task index, $U_{ij}$ is the utility of choosing task $j$ for robot $i$, $w_j$ is the weight of the $j$-th task, and $\alpha$ is a non-nagetive integer. The goal is to determine $\alpha_{ij}$ for all robot and task pairs that maximizes the weighted sum of utility in Equation (\ref{eqn:single-task}). Several challenges studied for single-task robot and single-robot task include interrelated tasks and resource contention among robots~\citep{nam2015assignment}.

There are cases where multiple robots can allocate the same task. For example, each robot can execute at most a single task while some tasks can be shared (single-task robot;~\cite{gerkey2004formal}), and some tasks can be achieved by multiple robots (multi-robot task;~\cite{gerkey2004formal}).
A coalition-formation game is adopted by \citet{jang2018anonymous} to form subteams of robots where the selfish robots seek to increase their reward by forming a coalition. They provide a decentralized algorithm requiring a strongly connected communication graph and prove a suboptimal Nash stable partition under certain conditions.

The online assignment problem is solving task allocation repeatedly to ensure dynamic task allocations on the fly in many robots and tasks. \citet{liu2012large} study online assignment problem in terms of single-task assignment that dynamically mixes centralized and decentralized approaches to generate efficient and robust allocation.

\subsection{Multi-task assignment}\label{subsec:multi_task}
Often, each robot from a robot team must be able to deal with more than a single task, particularly when the number of tasks (\eg, monitoring sites or targets of interest) is larger than that of robots. Robots can determine their assigned tasks either a priori for stationary tasks, or online for tasks that are dynamic in space and time.

Multi-robot multi-target assignment is fundamentally more challenging than conventional multi-task assignment among immobile agents because planning robot trajectories is also involved. It is undesirable in terms of time and energy that one robot travels a long distance to reach a task where another robot is nearby. Thus, distributing a fair amount of traveling distance and a fair number of tasks is critical. Communication challenges, such as limited communication range, limited bandwidth, and communication delay, can bring additional complexity. 

Combinatorial optimization techniques, such as mixed-integer linear programming and mixed-integer quadratic programming, are often employed to solve the assignment problem between a set of robots and a set of tasks. However, most multi-task assignment problems are NP-hard~\citep{vazirani2013approximation}, increasing the computational complexity dramatically as the number of robots and tasks increases. Algorithmic efforts have been mostly focused on alleviating the computational complexity by developing approximation algorithms~\citep{vazirani2013approximation} or distributed algorithms~\citep{lynch1996distributed}.

To avoid a single central processor that manages all information and computations,  \emph{auction algorithm}~\citep{bertsekas1988auction} has been popular as distributed algorithms, which works for both single and multi-task assignment. The idea of the auction algorithm originates from market-based coordination, where robots are considered self-interested participants and bid for tasks. The value of a task is computed by the difference between a reward and a price (\eg, distance robot and task location) of achieving the task. An auctioneer determines an appropriate selection of bids among robots based on computed task values. The auction algorithm achieves globally efficient solutions from selfish robots by maximizing their rewards and minimizing others' rewards~\citep{zlot2006auction}. Variants of the auction algorithm have been proposed to include additional features and show theoretical guarantees, including exploiting local information~\citep{zavlanos2008distributed}, a conflict-free assignment guarantee~\citep{choi2009consensus}, and optimal assignments~\citep{liu2013optimal}.

In multi-task assignment literature, various algorithmic challenges motivated by practical applications have been proposed. Examples, but not limited to, include tasks exhibiting different levels of priority~\citep{notomista2019optimal}, balancing the workload among robots to reduce idle robots~\citep{schwarzrock2018solving}, scheduled tasks with time window~\citep{nunes2015multi,suslova2020multi}, and disjoint groups of tasks~\citep{luo2014provably}.

\section{What Samples to Collect}~\label{sec:what}
There are often requirements for physically collecting ex-situ samples that are spread in the environment, especially when samples are needed to be analyzed in the lab. Due to the limited load capacity (\ie, budget) of a robot, only a limited number of samples can be gathered by a robot in a single deployment. 
\NEW{
The general formulation of this problem can be written as $\max R(S)\ \textrm{s.t.}\ |S|\le B$, where $R$ is a scalar function quantifying the quality of samples, $S$ is a set of collected samples, and $B$ denotes a given budget, which represents the maximum capacity of samples that can be collected.
}

Suppose all the sample information, such as the location and quality, is revealed to the robot a priori. In this case, the optimal strategy of collecting samples can be constructed, maximizing a reward or utility given a limited budget. However, in reality, partial or no information is given in most cases, which makes finding an optimal offline solution often infeasible. Although a sample currently available to be collected exhibits a good quality of interest, the robot may still want to give up on that sample and move on to the next sample, hoping that samples at a later time have better quality. Given these challenges, the robot must be able to determine what samples to collect among the stream of observed samples in exploring unknown and stochastic environments.

The \emph{hiring} (or) \emph{secretary problem}~\citep{freeman1983secretary,ferguson1989solved} involving the optimal stopping theory~\citep{shiryaev2007optimal} addresses a relevant problem that can be related to the question of what samples to collect. The problem formulation is as follows. Given $n$ number of secretary applicants of unknown quality, the interviewer interviews one applicant at a time, evaluates the quality of the applicant, and determines whether to hire the applicant. If the interviewer decides to hire the current applicant, the process ends; otherwise, the next applicant comes in. The process repeats until one applicant is hired. The rejected applicant cannot be considered anymore, so the interviewer makes irrevocable decisions. \citet{das2015data} present a data-driven sampling strategy that minimizes cumulative regret for batches of plankton samples using a combination of Bayesian optimization in the bandit setting, with online best-choice with hiring algorithm.
\citet{flaspohler2018near} propose irrevocable sample selection for a periodic data stream.
\citet{manjanna2018heterogeneous} consider two robotic boats for exploring the lake and sampling water samples using the Gaussian process-based modeling. They propose a variant of the secretary problem for finding the sampling locations, but their version allows the robot to visit previous sampling locations.

% Open problems. Heterogeneous samples.

\section{When to Collect Samples}~\label{sec:when}
When environmental monitoring involves multiple complex objectives, such as avoiding obstacles, reducing state uncertainty, and periodically collecting samples, it becomes challenging to determine the timing of the action of collecting samples from an environment.
Nonetheless, the question of when to collect samples is less explored than other decision-making problems in the literature.  

Since deciding when to collect samples is related to temporal reasoning, previous work has studied it in formal verification, specifically \emph{linear temporal logic} (LTL). In LTL, a user defines task specifications, such as eventually or infinitely often holding an event (\ie, LTL formulas) along with many others, and synthesizes a control policy that satisfies those specifications. As such, deciding when to collect samples can be encoded as one of the specifications along with other task specifications to determine the sample collection timing implicitly.

\citet{lu2018mobile} use the LTL to enforce a team of robots to infinitely often visit a partitioned set of environments.
\citet{nilsson2018toward} and~\citet{leahy2019control} employ the LTL for aerial-ground robots to collect samples from uncertain environments collaboratively.
For interested readers, refer to the survey papers on formal specifications in robotics~\citep{kress2018synthesis,luckcuck2019formal}.

The intermittent deployment problem proposed by~\citet{liu2019submodular} addresses the challenge of dependency between current and future monitoring to determine when to deploy a team of robots for monitoring (\eg, collecting samples). This challenge arises from a scenario where current monitoring influences underlying environmental dynamics.

\section{How to Learn Environment}~\label{subsec:how}
Generally, environment representation is not given a priori; thus, we must learn task-specific representation by gathering relevant data in the field. In Chapter~\ref{chap:env}, we have introduced numerous representations and their properties. This section focuses on existing environmental monitoring methods, leveraging recent machine learning advances. It is particularly fascinating to review machine learning techniques due to the recent popularity of deep learning-based approaches. An extensive summary of machine learning applied for data-driven geoscience can be seen in~\citet{bergen2019machine}. We leave out the Gaussian process-based methods in this section (\ie, learning hyperparameters in a data-driven way) as we present numerous studies in Chapter~\ref{chap:env} and Section~\ref{sec:where}.

\subsection{(Un)supervised learning}\label{subsec:supervised}
Following the great success of replacing traditional machine learning approaches~\citep{nuske2011yield} with deep neural networks (DNNs) in the computer vision community, most efforts have been spent appropriately adopting deep learning-based vision capabilities in environmental monitoring tasks. This \emph{supervised learning} framework learns to predict attributes of an object of interest (\eg, location, size, and color) from an image input, requiring human annotation to create a training dataset.

Numerous studies in agricultural robotics~\citep{r2018research,vougioukas2019agricultural,sparrow2021robots,oliveira2021advances,basiri2022survey} and forestry~\citep{oliveira2021forest} have applied convolutional neural networks (CNNs) to detect objects of interest, such as weed classification~\citep{sa2017weednet}, tree diameter estimation~\citep{chen2020sloam}, and corn stand counting~\citep{kayacan2018embedded}.
\citet{hani2020comparative} conduct a comparison study between the deep learning method and classical method (\ie, Gaussian mixture model-based approach) on apple detection and counting algorithms in orchards. They show that the deep learning method outperforms the classical technique on counting while the classical method still performs well on detection. \citet{chen2020geomorphological} use CNNs to detect and segment individual rocks on a rocky fault scarp, building a semantic map of rocks to study the formation and development of rocky fault scarp processes.  

Neural networks have also been applied to improve planning performance. 
\citet{saroya2020online} use CNNs to learn topological features in mining tunnels and caves to inform frontiers for robots, enhancing subterranean exploration.
\citet{li2021attention} propose the convolutional neural network-recurrent neural network to learn a multivariate spatiotemporal environment. They also incorporate the attention mechanism to detect parameterwise dependencies and spatial correlations, driving active sensing.

\emph{Weakly supervised learning} refers to supervised learning with noisy labeled data.
\citet{koreitem2020one} learn a visual similarity operator via a weakly supervised method to guide visual navigation. 

The \emph{self-organizing map}~\citep{faigl2016self,best2018online,jayaratne2019unsupervised,best2020decentralised} has been developed to learn a lower-dimensional representation of an input space without its violating topological properties. This \emph{unsupervised learning} procedure is used for active perception and data collection.

\subsection{Reinforcement learning}\label{subsec:reinforce}
Due to the burst of advances in \emph{reinforcement learning} (RL), numerous recent studies have employed RL to learn a policy for robots via online interaction with environments. RL algorithms developed recently show promising performances in sample efficiency and generalization, known challenges in RL, motivating researchers in environmental monitoring to adopt for their applications.

Several previous work has extended existing RL algorithms to tackle environmental monitoring-specific challenges. 
\citet{martin2017extending} address heteroscedastic marine environments where noise varies over states and actions, as opposed to environments having homogeneous noise in general RL problems, and propose a model-based algorithm.
In their subsequent work~\citep{martin2018sparse}, they propose a model-free algorithm for navigation in marine environments by reducing temporal difference updates to Gaussian process regression for data efficiency.
\citet{chung2015learning} develop resource-constrained exploration that is adaptive and directed. They use the Gaussian process regression to model the value function and propose a Gaussian process-based state–action–reward–state–action (SARSA) algorithm. To generate non-myopic action sequences, the objective function is involved with resource weights and a future information gain.
\citet{chen2019deep} design a deep RL tree to find near-optimal sampling locations by computing maximum information gain from the spatiotemporal field.
\citet{notomista2022multi} study RL to learn a coordinated-control policy for multiple robots based on a control-theoretic measure of the information (\ie, a norm of the constructability Gramian).

The RL framework has been adopted for various environmental monitoring applications. We present several studies demonstrating how environmental monitoring tasks can benefit from RL.
\citet{zhi2019learning} construct a 2D map by obtaining semantic segmentation obtained by visual images, which then becomes an input to the deep RL framework to determine the robot's action.
\citet{niroui2019deep} introduce deep RL to the conventional frontier-based exploration to handle high-dimensional state spaces. Their deep RL takes the map, robot, and frontier locations as input and yields a goal location as output. They adopt asynchronous advantage actor-critic as a learning tool instead of deep q-network due to its slow convergence rate.
\citet{doroodgar2014learning} propose hierarchical RL that can learn and make decisions with human supervision given multiple tasks. In their subsequent work~\citep{hong2018investigating}, they extend to the control of multi-robot teams so that the operator reduces interaction effort while improving task performance.
\citet{chen2019adaptive} employ RL to a known environment coverage task to reduce application-specific customization. 
\citet{ruckin2022adaptive} propose a Monte Carlo tree search with a convolutional neural network for informative path planning in unknown environments. 

In urban search and rescue, (hierarchical) RL is adopted to learn to explore unknown environments while allocating rescue tasks to multiple robots~\citep{doroodgar2014learning,hong2018investigating,niroui2019deep}.

\subsection{Other learning methods}\label{subsec:other_learning}
\emph{Graph neural network} (GNN;~\cite{wu2020comprehensive}) is designed to perform inference from graph-structured data, leveraging the notion of \emph{equivariance} for efficient learning. GNN is particularly useful for multi-robot systems as a graph can naturally represent the communication network among robots.
\citet{tolstaya2020multi} exploit GNN to learn the spatial equivariance of the multi-robot coverage task so that it can generalize to larger maps and larger teams.
\citet{gosrich2022coverage} focus on designing a GNN-based decentralized control policy addressing limited sensing range within a team of robots.

\citet{tompkins2020online} consider \emph{domain adaptation} of occupancy map models leveraging the observation that real-world structures contain similar geometric features. Their optimal transport-driven method learns to transfer parameters from domain to domain, static to dynamic environments, and simulation to real-world.

\NEW{
DNNs have the capacity to learn hierarchical features and establish global connections, allowing them to capture both local and global patterns in data. In contrast, Gaussian Markov random fields (GMRFs) operate under conditional independence assumptions, resulting in a sparser, more localized connectivity pattern~\citep{xu2016bayesian,kreuzer2018learning,duecker2021embedded}. GMRFs excel in modeling such data due to their localized connections. However, when environmental processes are influenced by global phenomena or when the dataset is extensive and varied, the flexibility and capacity of DNNs to learn from the data become crucial. The choice between DNNs and GMRFs should be guided by the specific objectives of the modeling effort. If the goal is to understand local spatial relationships or to perform spatial interpolation, GMRFs provide an efficient and interpretable option. For tasks that require capturing more complex patterns or when working with large and diverse datasets, DNNs might be more appropriate.
}

\section{Who to Communicate}~\label{sec:who}
Often, the field of interest is extensive and cannot be monitored by a single robot. Thus, we deploy multiple robots to aggregate local information obtained by a single robot to estimate global information. 
When deploying multiple robots for environmental monitoring, the robots must decide who to communicate due to the limited communication capability. As communication challenge in multi-robot systems has long been researched in robotics (see the survey papers by~\cite{yan2013survey,rizk2019cooperative,gielis2022critical}), we introduce environmental monitoring-related studies in this section.

Communication is one of the major bottlenecks, especially for field robots to deploy, as a stable WiFi connection is hard to achieve in practice due to the jerky motion of robots, mechanical issues, and WiFi interference. Thus, existing methods assuming perfect communication (\ie, centralized communication or all-to-all communication) may not be readily applicable to real-world deployment. We need a mechanism to enhance robot coordination, such as a communication map~\citep{li2019multi}.

Distributed \emph{consensus} is one of the frequently-used coordinating methods whose objective is to reach an agreed value among multiple robots. Measurements taken by neighboring robots are averaged and shared through communication. Analysis of convergence is critical in consensus-based algorithms. 
\citet{lynch2008decentralized} allow for heterogeneity and unknown and varying communication network among robots.
\citet{aragues2012distributed} propose a dynamic consensus method for map merging applications.
\citet{tamjidi2019efficient} improve a consensus filter for a dynamically changing network of robots by separating correlated information from uncorrelated one to achieve robustness to network failure. 
\citet{jang2020multi} propose a consensus-based distributed Gaussian process algorithm.

% \cite{capitan2013decentralized} design an \emph{auction}-based partially observable Markov decision process (POMDP) to reduce the communication burden while sharing policies within a team of robots. They demonstrate their method in an environmental monitoring scenario.
% \cite{varadharajan2020soul} develop a data sharing mechanism based on auction.

% \subsection{Topology control}\label{subsec:topology}
\emph{Topology control} allows robots to actively determine which communication link to enable from a set of available links under limited communication and sensing constraints.
\citet{williams2013constrained} propose a distributed interaction control scheme considering a set of topological constraints. Their scheme allows robots to actively retain and reject communication links from their neighbors, which is discrete switching, and to control based on attractive and repulsive potentials fields. Aggregation and dispersion with node degree constraints are verified with stability analysis.
\citet{mukherjee2020optimal} study optimal topology selection when asymmetric interactions among robots exist.

Multi-robot coordination requiring all-time connectivity is undesirable as this constraint may deteriorate the task performance of multi-robot systems.
Breaking the connectivity for some time may yield more flexible planning, called \emph{periodic connectivity} in the literature.
\citet{hollinger2012multirobot} develop an online algorithm scaling linearly with the number of robots and allowing for periodic connectivity constraints. They also show inapproximability results for connectivity-constrained planning.
\citet{wu2012robust} propose a switching strategy between individual and cooperative exploration to find a local minimum of an unknown field. Robots switch to cooperative exploration when converging to a local minimum at a satisfying rate is not detected and to individual exploration when the signal-to-noise ratio is improved.
\citet{banfi2018multirobot} study multi-robot reconnection planning on graphs. Their algorithm consists of two phases: optimizing the environmental monitoring objective and reconnection planning to regain global multi-hop connectivity.

While periodic connectivity enforces fixed-interval connectivity, \emph{intermittent interaction} relaxes this fixed-interval constraint and asks an algorithm to determine when to connect.
\citet{kantaros2019temporal} investigate a scenario where multiple robots conduct an information-gathering task without requiring the connectivity condition at all times. Their approach is based on linear temporal logic (LTL). They propose a global LTL that ensures intermittent connectivity among groups of robots and a local LTL that achieves a local task of a robot, such as the monitoring task. Their algorithm is distributed, online, and asynchronous.
\citet{heintzman2020multi} address an interaction planning problem, avoiding obstacles and respecting the constraints on random position samples. Their method is greedy and yields a theoretical guarantee using submodular objective function and matroid constraint.

% \subsection{Semi-Autonomous Approach}~\label{subsec:semi}
% Human supervision/intervention.
% Human-robot collaboration for sea inspection by Lindemuth~\etal~\citep{lindemuth2011sea}

% A hierarchical RL framework proposed by works~\citep{doroodgar2014learning,hong2018investigating} aims at semi-autonomous search and rescue by human operator.
% Shkurti~\etal~\citep{shkurti2012multi} presented a heterogeneous multi-robot system with off-site scientists for marine monitoring in order to find interesting visual footage. They discussed about the complexities of multi-robot experiments.
% Girdhar~\etal~\citep{girdhar2019streaming} studied scientist in the loop, co-robotic exploration in the deep sea.

%%%%%%%%%%

%%%%%%%%%%
\chapter{Conclusions and the Future Work}\label{chap:conc}
In this article, we survey algorithms for efficient sampling for environmental processes. First we discuss robust representations of the environment, followed by a listing of various tasks we are interested in carrying out on these environments, and finally, algorithms to execute these tasks.

Environmental monitoring involves the development and implementation of comprehensive autonomous systems, incorporating key components such as perception, state estimation, planning, and control. In addition to improving the performance of algorithms within the taxonomy to create reliable and successful systems for deployment in the real world, there are several promising future directions that have yet to receive sufficient research attention. We conclude the survey by introducing these future directions.

\paragraph{Heterogeneity}
Environmental monitoring presents inherently complex challenges, even when considering a single type of robot and a single type of sensor. It is evident that a heterogeneous team of robots or the utilization of heterogeneous sensor modalities within a single robot holds great potential for enhancing monitoring performance and addressing challenges that homogeneous teams or sensors may struggle with. For instance, the collaboration between aerial robots and ground robots can leverage the agility of aerial robots and the close perception and physical interaction capabilities of ground robots, resulting in synergistic benefits. Agriculture robots equipped with a combination of camera sensors and heat sensors have the potential to outperform robots equipped with either sensor alone. However, to fully harness the advantages of heterogeneity in environmental monitoring, there is a pressing need for a principled approach to merging different environment representations and addressing algorithmic considerations, which currently remains an open problem.

\paragraph{Environment Manipulation} In environmental monitoring research, the focus has primarily been on robots serving as passive observers or collectors of samples, with limited consideration given to their active interaction with the environment. However, in the real world, there are numerous scenarios where manipulability would offer significant benefits. For instance, in hazardous environmental monitoring, the ability to push and manipulate hazardous materials would optimize the collection of a larger amount in a single deployment. Moreover, manipulability equates to increased exploration capabilities for robots. Leveraging techniques developed in manipulation planning and control can not only address new challenges in environmental monitoring but also tackle more complex monitoring tasks. By embracing the potential of robot manipulability, the field of environmental monitoring can advance and effectively tackle a wider range of monitoring objectives.

\paragraph{Datasets, Simulations, and Models}
Following the trends in the vision and language communities, several benchmark simulations have recently been developed in robotics. These simulations serve the purpose of facilitating performance comparison among various methods and data collection for learning or pretraining. A crucial aspect of simulations is to minimize the sim-to-real gap, ensuring that running a robot in simulation closely resembles its behavior in the real world. However, there is limited research that applies the same level of effort to field robotics, specifically in the domain of environmental monitoring. Conducting robot experiments in real-world environments is significantly more challenging and resource-intensive, which justifies the need for building realistic and reliable simulations. Additionally, as algorithmic efforts in environmental monitoring are diverse and not easily comparable to other approaches, the development of such simulations can help converge towards promising approaches in the community and boost collaborative efforts.

Environmental monitoring typically occurs in unstructured, dynamic, and uncertain physical environments. One of its primary objectives is to gain a deeper understanding of scientific phenomena, which can significantly enhance the achievement of monitoring goals compared to situations where the environmental dynamics are poorly understood. Surface vehicles and underwater robots, for instance, can greatly benefit from understanding flow currents to accomplish their objectives more effectively. Similarly, monitoring natural disasters such as fires would be facilitated if the underlying physical laws governing fire behavior can be revealed and utilized. While some efforts have been made in this direction, there are numerous valuable techniques and methods from scientific fields that can be leveraged to develop physics-informed perception, planning, and control approaches. By harnessing these approaches, environmental monitoring tasks can be tackled with greater efficacy and efficiency.

% 1. Heterogeneity:
% \cite{pretto2020building}: aerial-ground robots for precision farming.

% 2. Continuous environment representation: how to abstract, connection to SLAM where the map is finely discretized (micro level).

% No consensus on environment representation. The choice of algorithms depends on the representation chosen.

% 3. Robots as actors that affect the state of the environment (e.g. by manipulating objects).

% \begin{acknowledgements}

% \end{acknowledgements}

%BACKMATTER SEE DOCUMENTATION
\backmatter  % references, restarts sample

\printbibliography

\end{document}